\documentclass[10pt,twocolumn,letterpaper]{article}
\usepackage[nopdf]{epstopdf}

\usepackage[pagenumbers]{cvpr}

\usepackage{amsfonts}
\usepackage{graphicx}
\usepackage{pifont}
\usepackage{tabularx}
\usepackage{multicol}
\usepackage{multirow}
\usepackage{threeparttable}
\usepackage{caption}
\usepackage{xcolor}
\usepackage{makecell}
\usepackage{tcolorbox}
\usepackage{booktabs}
\usepackage{bm}
\usepackage{adjustbox}
\usepackage{placeins}

\definecolor{cvprblue}{rgb}{0.21,0.49,0.74}
\usepackage[pagebackref,breaklinks,colorlinks,allcolors=cvprblue]{hyperref}

\title{From Concept Erasure to Style Purification: Contrastive Eigenbases for Artist Style Protection}

\author{Tong Zhang$^{1}$,\; 
	Ru Zhang$^{1,*}$,\; 
	Jianyi Liu$^{1}$,\; 
	\\ 
	\normalsize $^{1}$ Beijing University of Posts and Telecommunications, School of Cyberspace Security, Beijing, 100876,\;\\ 
	{\tt\small *Corresponding author}}

\begin{document}

\maketitle

\begin{abstract}
Text-to-image diffusion models can reproduce the visual styles of specific artists at extremely low cost, making unauthorized artist style mimicry an important copyright and deployment safety concern for platforms. Existing model side protection methods typically follow the paradigm of ordinary concept erasure, focusing on more aggressive deletion or redirection of target styles. However, we find that for artist styles, the fundamental issue is not insufficient erasure strength, but the mismatch between artist styles and the ordinary concept erasure paradigm. Through causal intervention analysis, we observe that artist styles do not correspond to relatively compact and localized editable semantic units like ordinary object concepts. This characteristic makes it difficult for sparse editing and fixed retain lists to simultaneously achieve target suppression and generation utility preservation. Based on this observation, we reformulate artist style protection from concept erasure to style purification, where the target style expression is suppressed during inference while preserving the requested content and visual structure. We propose CAPE (Contrastive Artist Style Purification with Eigenbases), a training-free framework for protecting against artist style mimicry. CAPE first constructs contrastive triplets around the target request and formulates style direction estimation as a solvable generalized eigenvalue problem to capture style related directions that remain stable across content variations and are less affected by shared content. During inference, CAPE employs the Adaptive Suppression Controller to assign suppression strengths to different tokens based on Q, K, and V responses, and performs target style suppression on the K and V paths of self-attention. Experimental results show that CAPE effectively weakens target artist characteristics, including brushstrokes, textures, and local color processing, while better preserving major semantic entities, scene composition, and visual structures.
\end{abstract}    
\section{Introduction}
\label{sec: introduction}

Text-to-image diffusion models have significantly lowered the barrier to mimicking specific artist styles. Users can directly specify artist names or indirectly trigger similar visual expressions through descriptions of brushstrokes, textures, and other characteristics, and apply them to new objects, scenes, and narrative contents. Different from copying a specific existing artwork, artist style mimicry is closer to a conditional re-rendering process. The model uses the visual patterns of the protected artist to re-render the requested content according to the user's content requirements. This capability brings new possibilities for creation, but also requires platforms to address unauthorized style mimicry and related deployment risks \cite{shan2023glaze,somepalli2024measuring}.

Existing artist style protection methods include data perturbation, prompt filtering, and model intervention. Among model intervention methods, concept erasure has become one of the primary approaches for suppressing target styles. Data perturbation adds imperceptible perturbations to publicly available artworks of artists \cite{shan2023glaze,shan2024nightshade,ahn2025nearly}, reducing the ability of models to learn or reproduce their styles from training data. Prompt filtering detects, removes, or rejects artist names and related keywords to block explicit requests. However, safety mechanisms relying on surface text matching are vulnerable to implicit rewriting or adversarial prompts \cite{yang2024sneakyprompt,tsai2024ring}. Model intervention includes fine tuning, optimization-based weight editing, and inference time intervention \cite{gandikota2023erasing,lu2024mace,huang2024receler,wang2025precise,li2026speed}. These methods usually redirect target representations toward null-text or neutral anchors, or directly suppress parameters, activations, and attention features associated with target concepts. Although they can reduce target style expression, they may introduce uncontrollable generation utility degradation. Therefore, model side protection is commonly organized around two objectives: increasing target suppression and reducing collateral degradation. Artist style protection is usually treated as an erasure strength and precision problem.

This erasure behavior is not only determined by the intervention strategy itself, but is also closely related to the internal representation characteristics of different concepts \cite{basu2024localizing,somepalli2024measuring}. Object concepts usually appear as visual entities with relatively clear semantic boundaries, whereas artist styles are expressed through rendering patterns such as textures, brushstrokes, and contour processing. Therefore, we investigate a more fundamental question: whether artist styles are suitable targets for ordinary concept erasure.

We conduct two types of analyses across different concept categories, including the distribution of erasure causal contributions among model components and the extent of collateral effects introduced by erasure on non target concepts. We observe that artist styles do not correspond to compact and localized editable semantic units like ordinary object concepts. The contribution mass of artist styles is more difficult to cover using a small number of highly contributing components, and cumulative suppression increases more slowly under sparse intervention. Meanwhile, compared with other concept categories, the collateral effects of artist styles are less concentrated within local semantic neighborhoods and are more likely to extend to distant non target concepts. These observations indicate that artist styles are not suitable targets for relying solely on sparse model editing and limited fixed retain lists.

Based on this observation, we no longer treat artist style protection as an ordinary target erasure problem, but reformulate it as style purification. Unlike the erasure paradigm that modifies model parameters, style purification intervenes in the generation process only when the target style is requested, reducing collateral effects introduced by modifying shared internal representations while preserving the requested content and normal generation utility as much as possible. Meanwhile, different from inference time intervention methods that redirect the generation process, style purification does not require mapping the target style to predefined replacement representations, but directly suppresses responses related to target style expression under the current prompt condition. Therefore, the key challenge is no longer how to maximally suppress internal representations corresponding to the target style, but how to identify target style related responses in the current generation process and selectively suppress them.

To address this, we propose CAPE (Contrastive Artist Style Purification with Eigenbases), a training free framework for artist style protection. We construct contrastive triplets from mimicry prompts to capture style signals that remain stable across content variations, and formulate this process as a solvable generalized eigenvalue problem. Subsequently, we use the estimated style signal to perform projection suppression on the K and V paths of self attention, and introduce the Adaptive Suppression Controller to adaptively adjust the suppression strength according to the responses of the style signal across different branches, addressing the non uniform response of target styles across spatial tokens. CAPE requires neither additional training nor iterative optimization, and does not rely on predefined replacement styles. It only requires a single feature extraction process and generalized eigenvalue solving, resulting in lower inference and computational costs. While suppressing target style expression, CAPE preserves the model’s general generation capability and the structure of the requested content. We systematically evaluate CAPE on multiple artist styles with diverse visual expression characteristics and compare it with representative concept erasure and inference-time intervention baselines. The results demonstrate CAPE's ability to balance style suppression and content preservation. CAPE weakens target artists’ brushstrokes, textures, and local color processing while causing fewer changes to major objects, background composition, and geometric structures. In summary, the main contributions of this paper are as follows:
\begin{itemize}
	\item Our analysis reveals that artist style does not exhibit a compact and localized editable semantic unit like ordinary object concepts from the perspective of erasure behavior, motivating the reformulation of artist style protection from concept erasure to style purification.
	\item We propose CAPE, which formulates inference-time style suppression as a generalized eigenvalue problem and introduces the Adaptive Suppression Controller to address the non uniformity of style expression.
	\item Cross style comparisons and systematic ablation studies demonstrate the effectiveness of CAPE in balancing target style suppression with content and structural preservation.
\end{itemize}
\section{Related Work}
\label{sec: related work}

\subsection{Concept erasure}
Concept erasure aims to prevent text-to-image models from generating a specified concept while preserving their ability to generate unrelated content. Early representative work ESD \cite{gandikota2023erasing} modifies model parameters using negative guidance to permanently erase target concepts. Subsequent approaches have improved scalability, robustness, and non-target preservation from different perspectives. MACE \cite{lu2024mace} combines closed-form refinement with LoRA fine-tuning to support mass concept erasure, while SPM \cite{lyu2024one} introduces lightweight specific adapters for modular and transferable erasure. Receler \cite{huang2024receler} further formulates robustness against paraphrased or adversarial prompts and locality with respect to non-target concepts as central objectives of reliable concept erasure.

Subsequent studies have focused on reducing editing cost and collateral degradation. EraseDiff \cite{wu2025erasing} formulates erasure and model utility preservation as a constrained optimization problem. AdaVD \cite{wang2025precise} performs training-free concept erasure through orthogonal decomposition in cross attention value space, while GLoCE \cite{lee2025localized} selectively activates lightweight low-rank modifications only for localized target concepts. SuMa \cite{nguyen2025suma} represents narrow concepts using subspaces and maps the target representation toward a reference subspace to balance erasure completeness and generation quality. Concept erasure has also been extended to flow-based Transformer architectures such as SD3 and FLUX \cite{gao2025eraseanything}.

Evaluation studies have shown that concept erasure can suppress non target concepts that are semantically or visually related to the target. EraseBench \cite{amara2025erasing} systematically evaluates such spillover degradation and reveals that non target preservation remains a central challenge for deployment. Mechanistic studies have also used causal analysis and tracing to localize visual attributes and identify editing relevant components within diffusion models \cite{basu2024localizing,zhang2025beyond}. These studies establish non target preservation and internal intervention structure as important dimensions for understanding concept erasure.

\subsection{Artist Style Protection against Mimicry}

From the perspective of protecting artist style based on training data, prior work has applied perturbations to released artworks to disrupt subsequent style learning or personalization. Glaze \cite{shan2023glaze} introduces imperceptible style cloaks that alter the apparent style learned from protected images and thereby disrupt unauthorized style mimicry. Related data poisoning methods such as Nightshade \cite{shan2024nightshade} corrupt the association between textual concepts and visual representations during model training. FastProtect \cite{ahn2025nearly} further reduces the computational cost of protective perturbations through perturbation pretraining and adaptive inference. These approaches act on the training signal before or during model adaptation, and therefore address a different stage of artist protection from direct control of the generation process.
\subsection{Style Representation and Inference Time Control}
 
A separate line of work investigates how style, appearance, and structure are represented and controlled inside diffusion models. Somepalli et al. \cite{somepalli2024measuring} introduce Contrastive Style Descriptors to characterize visual style through interactions among color, texture, shape, and other appearance factors, demonstrating that style similarity cannot be reduced to simple semantic similarity. Plug-and-Play Diffusion Features \cite{tumanyan2023plug} shows that intermediate diffusion features and self-attention representations contain spatial and structural information that can be directly manipulated during inference without model training. 

For artistic style control, Style Injection \cite{chung2024style} manipulates self-attention representations by transferring Key and Value features from a style image while preserving Query information to better retain content structure. ZipLoRA \cite{shah2024ziplora} instead combines separately learned subject and style LoRAs to jointly preserve subject identity and stylistic characteristics. B-LoRA \cite{frenkel2024implicit} exploits selected LoRA blocks to separate style and content factors for image stylization. These methods demonstrate that style appearance can be manipulated through internal representations. Their primary objective is to transfer, inject, or compose a selected style as part of the desired output appearance. The setting considered in this work instead focuses on suppressing a specified target style while retaining the requested content.

\section{Why Artist Style Erasure Differs from Object Erasure}

Existing concept erasure methods usually treat artist style as an ordinary concept. However, these two types of concepts exhibit different patterns of expression. Objects usually appear as visual entities with relatively clear semantic boundaries, whereas artist style affects content expression through the combined effects of texture, brushstrokes, color, and contour processing. To analyze whether such differences in expression affect erasure behavior, we investigate the erasure behavior of artist styles and other object concepts from two perspectives, including the distribution of causal erasure contributions across model components and the scope of non target collateral effects caused by erasure.

\subsection{Distributed Causal Support of Erasure}
To estimate the causal contribution of model components to erasure, we use the original model $M_{ori}$ and the edited safe model $M_{edit}$. Starting from the original model $M_{ori}$, we replace only one single cross attention layer component $u$ with the corresponding component from $M_{edit}$ at each time, while keeping all other components unchanged (Figure \ref{figure1}), resulting in the intervention model $M_u$. For a concept $c$, we quantify the contribution of component $u$ to erasure using the difference between the generated image metrics before and after intervention as the target concept response score $E_{c,u}$:

\begin{figure}[h]
	\centering
	\includegraphics[width=1\columnwidth]{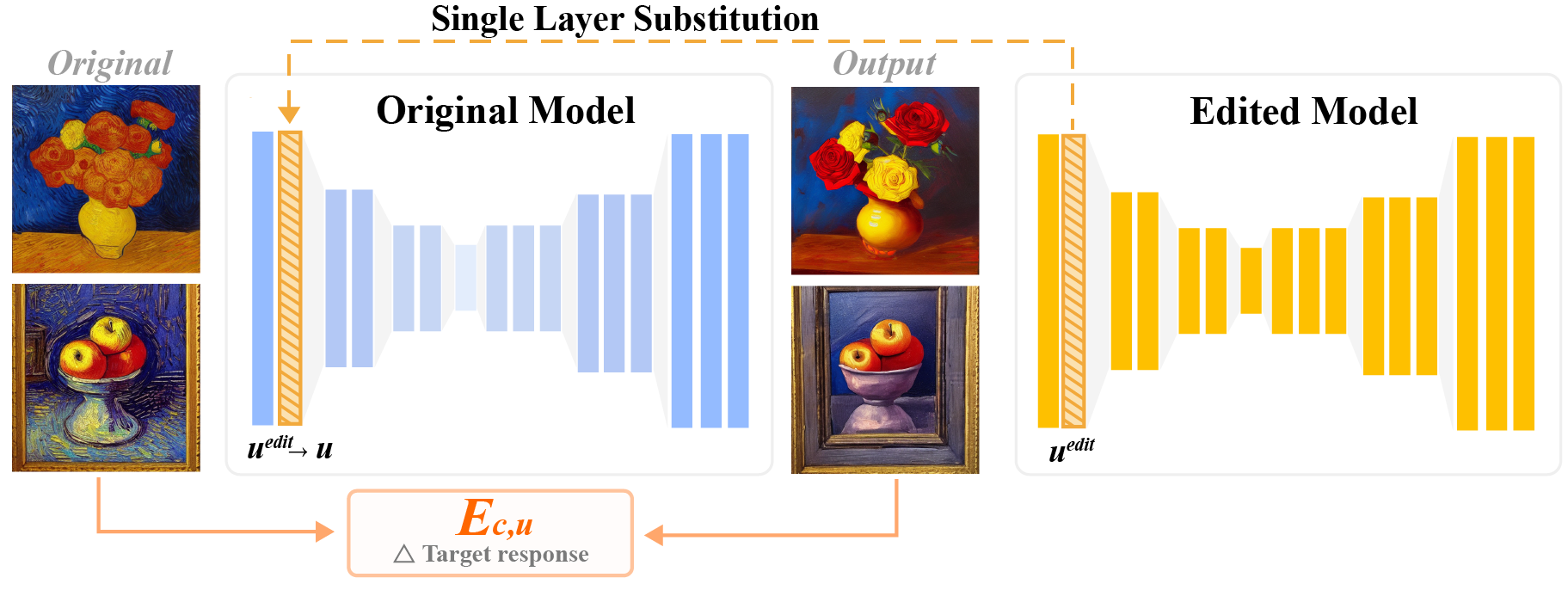} 
	\caption{Component level causal intervention for erasure analysis. Starting from the original model, we inject only one cross attention layer component from the erased checkpoint at each time while keeping the remaining components unchanged. The causal contribution of each component is quantified by evaluating the difference between the generated images before and after intervention.}
	\label{figure1}
\end{figure}
\begin{equation}
	\small
	E_{c,u}=R_{c}(M_{ori})-R_{c}(M_{u})
\end{equation}
where artist styles and object concepts use CLIP scores as the target concept response scores, while celebrity identities use Giphy Celebrity Detector(GCD) face recognition \cite{hasty2024giphy}. We normalize the contribution mass as:
\begin{equation}
	\small
	p_{c,u}=
	\frac{max(E_{c,u},0)}  
	{\sum_{j=1}^{U}max(E_{c,j},0)}
\end{equation}
We further measure the dispersion of causal erasure components for each concept through normalization:
\begin{equation}
	\small
	N(c)=
	\frac{1}  
	{U\sum_{u=1}^{U}p_{c,u}^{2}},U=16
\end{equation}
When $N(c)$ approaches 1, it indicates that multiple components have similar causal contributions, whereas a smaller value indicates that the erasure effect is concentrated in fewer units. We also use the Top-$k$ causal contribution mass metric $M_c(k)$, which represents the proportion of the total causal erasure contribution covered by the top-$k$ components with the highest contributions:
\begin{equation}
	\small
	M_c(k)=\sum_{u\in\operatorname{TopK}(c,k)}  
	p_{c,u}
\end{equation}
We conduct statistics on more than 70 successfully erased target concepts in each category, including artist styles, ordinary objects, and celebrity. The experimental results are shown in Figure \ref{figure2} and Table \ref{tab1}. The results show that artist style exhibits a higher component level causal erasure dispersion than other concept categories, while multiple Top-$k$ mass values are lower, indicating that the causal mass of artist style erasure is more difficult to cover using only a small number of high contribution components.

\begin{table}[t]
	\centering
	\caption{Distributed causal support of erasure across different concept categories. 
		Artist styles exhibit higher causal dispersion and lower Top-$k$ contribution mass compared with object and celebrity concepts.}
	\label{tab1}
	\footnotesize
	\setlength{\tabcolsep}{6pt}
	
	\begin{tabular*}{\columnwidth}{@{\extracolsep{\fill}}lccc}
		\toprule
		Metric
		& Artist style
		& Object
		& Celebrity \\
		\midrule
		
		$N(c)$ 
		& \textbf{0.702} 
		& 0.635 
		& 0.229 \\
		
		$M_c(1)$ 
		& \textbf{0.209} 
		& 0.262 
		& 0.724 \\
		
		$M_c(3)$ 
		& \textbf{0.484} 
		& 0.539 
		& 0.836 \\
		
		$M_c(5)$ 
		& \textbf{0.658} 
		& 0.700 
		& 0.893 \\
		
		\bottomrule
	\end{tabular*}
	
\end{table}

\begin{figure*}[h]
	\centering
	\includegraphics[width=1\textwidth]{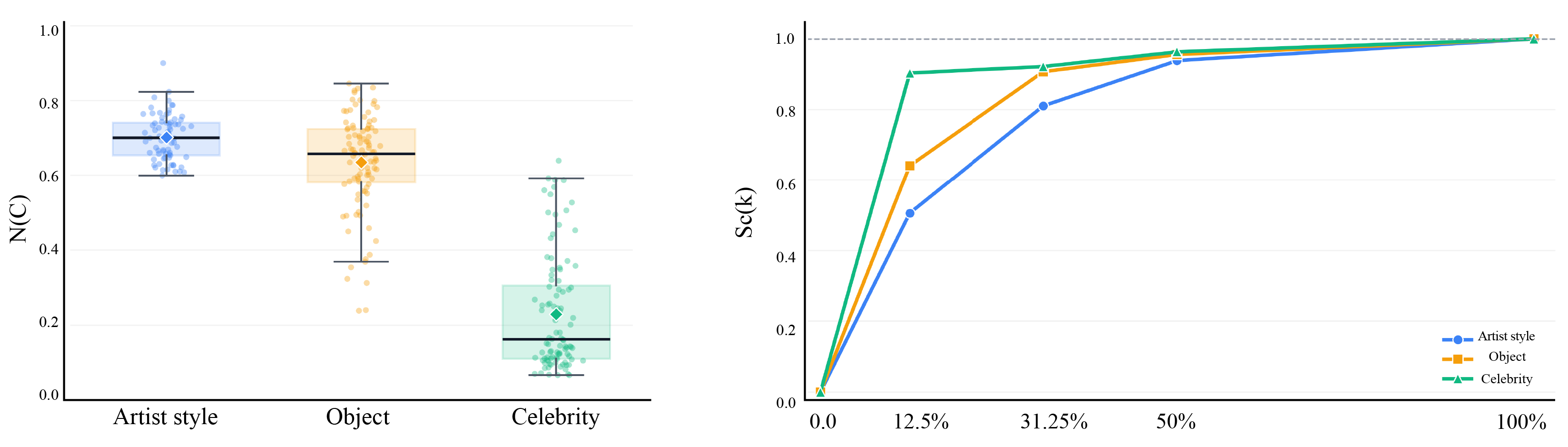} 
	\caption{(a) Box plots of causal component dispersion $N(c)$ for different concept categories. Compared with other concept categories, artist style exhibits more distributed effective causal support. (b) Cumulative target suppression under increasing cross attention coverage. Artist style suppression accumulates more slowly under sparse intervention and gradually approaches the full edit effect only with broader component coverage.}
	\label{figure2}
\end{figure*}

To verify whether this dispersed causal contribution directly affects the scope of effective intervention, we further evaluate interventions with different numbers of components. We replace components in descending order according to $E_{c,u}$ and divide by the response score of the complete editing model to measure how much of the full editing suppression effect is recovered by the current editing scheme:
\begin{equation}
	\small
S_c(k)=
\frac{R_c(0)-R_c(k)}  
{R_c(0)-R_c(16)}.  
\end{equation}
$R_c(k)$ denotes the target response after replacing the top-$k$ intervention units ranked by $E_{c,u}$. The experimental results (Figure \ref{figure2}) show that artist style suppression accumulates more slowly under sparse coverage conditions and only gradually converges after broader component intervention.

\subsection{Non-target Collateral Effect}
The above results indicate that artist style erasure requires broader component coverage. We further investigate whether this distributed intervention requirement is accompanied by a broader range of non target collateral effects. For each concept, we construct 24 non-target concepts, which are categorized into 8 near concepts, 8 middle concepts, and 8 far concepts according to semantic, contextual, scene related, and style or category proximity relationships. For each target concept, we use the original model and the model after target concept erasure to generate images for each non-target concept, and compare the changes between the generated images before and after erasure. We characterize collateral damage from two perspectives: content preservation and style related visual changes.

For the content dimension, we use CLIP cosine similarity to measure alignment changes \cite{radford2021learning} and DINOv2 global visual feature cosine distance \cite{oquab2023dinov2} to measure changes in overall visual representations. After normalization, the two measurements are combined with weighting to obtain $D_{i}^{\mathrm{content}}$.

For the style dimension, we use Contrastive Style Descriptors (CSD) to calculate the cosine distance before and after erasure, measuring the artistic style similarity change between generated images \cite{somepalli2024measuring}, and use the multi layer Gram representation of VGG-19 \cite{simonyan2014very,gatys2016image} to measure image style changes. Since these measurements have different scales, the results calculated by CSD and VGG-19 are independently combined with content collateral as two types of style representation collateral and content collateral to evaluate the final collateral damage:
\begin{equation}
	\small
 D_i^{content,m} = \frac{ D_{i}^{\mathrm{content}} + D_{i}^{m} }{2},\quad m\in{\{CSD,VGG\}}
\end{equation}
We calculate the proportion of total collateral effects located in the near concept and far concept sets, denoted as $D_{near}$ and $D_{far}$, as well as the proportion of the most severely affected Top 8 concepts that belong to the near concept and far concept sets, denoted as $D_{near}^{Top8}$ and $D_{far}^{Top8}$. The experimental results are shown in Table \ref{tab2}.
\begin{table}[t]
	\centering
	\caption{Non-target collateral footprint under CSD-based and VGG-based damage measurements.}
	\label{tab2}
	\setlength{\tabcolsep}{7pt}
	\footnotesize
	\begin{tabular*}{\columnwidth}{@{\extracolsep{\fill}}lcccc}
		\toprule
		\textbf{Metric}
		&
		\multicolumn{2}{c}{$D^{content,CSD}$}
		&
		\multicolumn{2}{c}{$D^{content,VGG}$}
		\\
		
		\cmidrule(lr){2-3}
		\cmidrule(lr){4-5}
		
		&
		Style
		&
		Object
		&
		Style
		&
		Object
		\\
		\midrule
		
		$D_{near}$
		&0.345
		&0.394
		&0.343
		&0.407
		\\
		
		$D_{far}$
		&0.325
		&0.294
		&0.337
		&0.274
		\\
		
		$D_{near}^{Top8}$
		&0.370
		&0.465
		&0.365
		&0.480
		\\
		
		$D_{far}^{Top8}$
		&0.323
		&0.248
		&0.328
		&0.233
		\\
		
		\bottomrule
	\end{tabular*}
	
\end{table}

Compared with object erasure, artist style erasure has a lower proportion of collateral damage concentrated in the near neighborhood, and the overlap between Top 8 high damage concepts and near concepts is also lower. In contrast, the collateral damage of object erasure is more concentrated around the local semantic neighborhood of the target concept. This observation is consistent with previous studies \cite{amara2025erasing,zhang2025beyond}, which show that object erasure has a stronger impact on semantically related concepts \cite{amara2025erasing}. Therefore, existing methods typically explicitly include neighboring concepts in the retain set to preserve non target semantics \cite{zhang2025beyond}. Our experiments further reveal the particularity of this issue in artist style erasure. For artist style concepts, collateral effects are less restricted to immediately adjacent semantic regions, making retain sets constructed solely based on limited and fixed neighborhoods more likely to have incomplete coverage of affected concepts.

Overall, from the perspective of erasure behavior, artist style does not correspond to a compact and localized editable semantic unit like ordinary object concepts. Its causal contributions are distributed across a broader range of model components, indicating that effective artist style suppression requires broader component coverage. This characteristic makes artist style erasure more difficult to achieve both effective style removal and non target protection through sparse intervention and limited retain sets. These observations expose a limitation of treating artist style as an ordinary erasure target and motivate a different formulation.
\section{Method}

Section 3 shows that artist style erasure has broader causal support than ordinary object erasure and leads to more dispersed non target collateral effects. Based on this observation, we argue that artist style should not be protected through erasure as an ordinary concept. Instead, we reformulate artist style protection as style purification. We propose CAPE (Contrastive Artist Style Purification with Eigenbases). Given a mimicry prompt $p_0=p(c_0,s^\star)$, CAPE suppresses the expression of the protected style $s^\star$ during inference while preserving the requested content $c_0$ and the general generation capability as much as possible.

\begin{figure*}[h] 
	\centering 
	\includegraphics[width=1\textwidth]{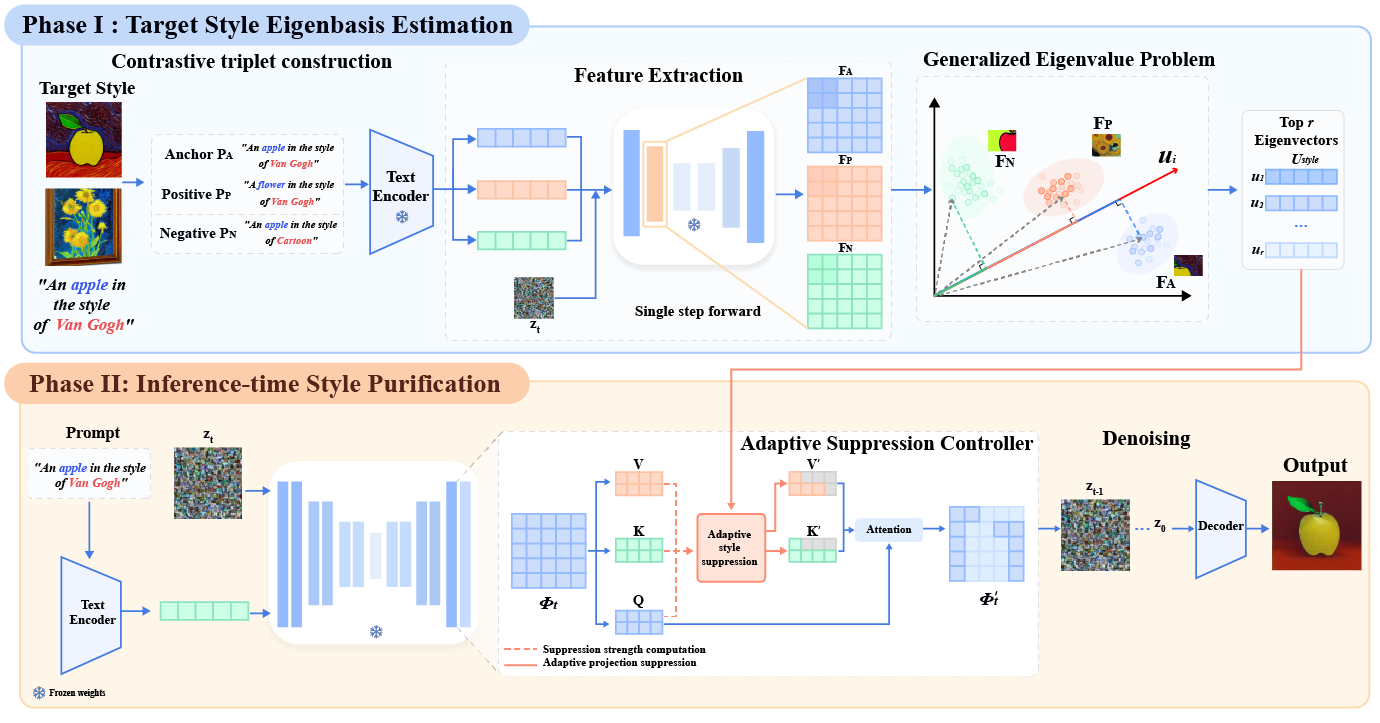} 
	\caption{Overview of CAPE. The first stage constructs contrastive triplets and estimates $U_{style}$ by solving a generalized eigenvalue problem. During generation, ASC estimates token level style concentration from Q, K, and V responses and adaptively suppresses the projected style associated responses in the K and V paths of self-attention.} 
	\label{figure3} 
\end{figure*}

CAPE consists of two stages. The first stage constructs contrastive triplets around the mimicry prompt and obtains an operational style signal by solving a generalized eigenvalue problem. The second stage estimates the style suppression strength of each token through the Adaptive Suppression Controller and suppresses style expression on the K and V paths. The overall framework is illustrated in Figure \ref{figure3}.

\subsection{Contrastive Triplets}
A single sample simultaneously contains style and content information \cite{somepalli2024measuring,frenkel2024implicit}. When features are extracted from an isolated sample, it is difficult to determine which features mainly reflect the target style and which are more related to specific content. Therefore, instead of directly defining the style direction from a single sample, we estimate it using contrastive triplets with controlled shared relationships. CAPE constructs contrastive triplets around the style mimicry prompt:
\begin{itemize}
	\item \textbf{Anchor} ($P_A=p(c_0,s^\star)$) : Contains the protected target style and the requested content.
	\item \textbf{Positive} ($P_P=p(c_m,s^\star)$) : Contains the same target style but different content.
	\item \textbf{Negative} ($P_N=p(c_0,s_l)$) : Contains the same content but a different style.
\end{itemize}
The $P_A$/$P_P$ relation provides positive evidence for directions that remain responsive to the target style across content changes, whereas the $P_A$/$P_N$ relation acts as a competing constraint that penalizes directions strongly supported by shared content. We feed the triplet prompt samples $P_A$, $P_P$, and $P_N$ into the model, extract feature maps $F_A$, $F_P$, and $F_N \in \mathbb{R}^{C \times H \times W}$ from the same steps and layers, and reshape them into patch matrices $F_A$, $F_P$, and $F_N \in \mathbb{R}^{N \times D}$, where $N$ is the number of patches and $D$ is the feature dimension of each patch.

\subsection{Eigenbasis Estimation}
In the triplet, the Anchor and Positive share the target artist style, and their shared features mainly reflect style information that remains stable across content variations. In contrast, the Anchor and Negative share the same content, and their shared features are more related to content information. We aim to identify style signal directions that remain stable across content changes. Such directions should exhibit strong responses between $F_A$ and $F_P$ while reducing the influence of shared responses between $F_A$ and $F_N$. We formulate this process as a Rayleigh Quotient optimization problem \cite{fisher1936use,weenink2003canonical}:
\begin{equation}
	\small
	\underset{u}{\operatorname*{arg\,max}} \ \mathcal{J}_s(u)= 
	\frac{u^\top (\Sigma_{AP} \Sigma_{PA}) u}{u^\top (\Sigma_{AA} + \lambda \Sigma_{AN} \Sigma_{NA}) u} 
\end{equation}
where $\Sigma_{AP}$ and $\Sigma_{AN}$ denote the covariance matrices computed from the corresponding centered feature pairs. The numerator $u^\top (\Sigma_{AP} \Sigma_{PA}) u$ serves as the style reward term, encouraging the projection direction to capture the shared style information between Anchor and Positive. The denominator consists of the Anchor self covariance $\Sigma_{AA}$ and the content penalty term $u^\top(\lambda \Sigma_{AN}\Sigma_{NA})u$. The former normalizes the projection variance of the Anchor to achieve scale invariance, while the latter penalizes directions supported by shared content information. Maximizing this Rayleigh Quotient obtains style related directions that remain stable across content variations while containing less content related interference. 

Since $\mathcal{J}_{s}(cu)=\mathcal{J}_{s}(u)$ holds for any nonzero constant $c$, maximizing $\mathcal{J}_{s}(u)$ is equivalent to maximizing $u^{\top} \left( \Sigma_{AP}\Sigma_{PA} \right) u$ under the constraint:
\begin{equation}
	\small
	u^{\top} \left( \Sigma_{AA} + \lambda\Sigma_{AN}\Sigma_{NA} + \epsilon I \right) u = 1
\end{equation}
We solve this problem using the Lagrange multiplier method and construct the Lagrangian function $\mathcal{L}(u,\rho)$:
\begin{equation}
	\small
	 \mathcal{L}(u,\rho) = u^{\top} \left( \Sigma_{AP}\Sigma_{PA} \right) u - \rho \left[ u^{\top} \left( \Sigma_{AA} + \lambda\Sigma_{AN}\Sigma_{NA} + \epsilon I \right) u - 1 \right]
\end{equation}
Setting the gradient of $\mathcal{L}$ with respect to $u$ to zero yields the first-order optimality condition:

\begin{equation}
	\small
	 \nabla_u\mathcal{L} = 2 \left( \Sigma_{AP}\Sigma_{PA} \right) u - 2\rho \left( \Sigma_{AA} + \lambda\Sigma_{AN}\Sigma_{NA} + \epsilon I \right) u = 0 
\end{equation}
This simplifies to the generalized eigenvalue problem:
\begin{equation}
	\small
	  \left( \Sigma_{AP}\Sigma_{PA} \right)u = \rho \left( \Sigma_{AA} + \lambda\Sigma_{AN}\Sigma_{NA} + \epsilon I \right)u
\end{equation}
where $\epsilon I$ is a regularization term introduced to alleviate ill conditioning caused by finite sample size and rank deficient feature covariance matrices, thereby improving numerical stability in solving the generalized eigenvalue problem. Solving the generalized eigenvalue problem produces a set of generalized eigenvalues:
\begin{equation}
	\small
	 \rho_1 \ge \rho_2 \ge \cdots \ge \rho_D
\end{equation}
Larger generalized eigenvalues correspond to directions with stronger shared responses between Anchor and Positive while being less affected by shared responses between Anchor and Negative. Artist style is usually expressed through multiple visual factors, such as color, texture, and brushstrokes, rather than concentrated in a single direction. Therefore, we select the directions corresponding to the largest $r$ generalized eigenvalues and apply QR decomposition to obtain an orthonormal basis for the subspace formed by these directions:
\begin{equation}
	\small
	 U_{\mathrm{style}} = [u_1,u_2,\ldots,u_r] \in \mathbb{R}^{D\times r} \quad U_{\mathrm{style}}^{\top}U_{\mathrm{style}}=I 
\end{equation}
Finally, $U_{style}$ is used for subsequent target style suppression. Through the AP and AN relationships in the triplets, it is biased toward target style signals that remain stable across content variations.

\subsection{Adaptive Suppression Controller}
After obtaining the operational style signal, directly applying uniform suppression to the entire latent representation may lead to excessive suppression of normal content structures, because the response of the target artist style during generation usually exhibits spatially uneven distribution. For example, in Van Gogh style generation, the swirling brushstrokes in the background sky often exhibit stronger style characteristics than static objects. Applying a fixed suppression strength may result in insufficient purification in highly style related regions while causing excessive intervention in regions with weak style responses. To address this issue, we design the Adaptive Suppression Controller (ASC), which assigns suppression strengths according to the target style concentration of each token. For the $i$-th token, we compute the projections of its $Q$, $K$, and $V$ representations onto the style subspace $U_{style}$ to quantify the style concentration:
\begin{equation}
	\small
	s_i^{t} = \| x_i^{t} U_{style} \|_2 \quad t \in\{Q,K,V\}
\end{equation}
A larger $s_i^{t}$ indicates that the token has a stronger response to the target style direction. To eliminate the scale differences among different components, we perform normalization within each sample:
\begin{equation}
	\small
	\bar{s}_i^{t}= \frac{s_i^{{t}}}{\max\limits_{1\le{j}\le{N}} (s_j^{t})+\delta} \quad t \in\{Q,K,V\}
\end{equation}
where $\delta$ is a constant used to avoid division by zero. We then obtain the overall style score of each image patch through weighted summation:
\begin{equation}
	\small
	S_i
	=
	\sum_{t\in\{Q,K,V\}}
	w_t\,\bar{s}_i^{t}
\end{equation}
where $w_q < w_k < w_v$ controls the relative contributions of $Q$, $K$, and $V$ to the style concentration. We further map $S_i$ into a smooth modulation factor $m_i$ through the Sigmoid function $\sigma$:
\begin{equation}
	\small
	m_i = \sigma(k \cdot (S_i - \tau))
\end{equation}
where $\tau$ is the activation threshold and $k$ controls the steepness of the intensity variation. Finally, the adaptive style suppression strength of the $i$-th token is defined as:
\begin{equation}
	\small
	\gamma_i = \alpha_{min} + (\alpha_{max} - \alpha_{min}) \cdot m_i
\end{equation}
ASC applies stronger suppression to tokens with higher target style responses, prioritizing regions that contribute more to the target artist style while applying weaker intervention to low response tokens, thereby reducing unnecessary content distortion.
Furthermore, we observe that $Q$, $K$, and $V$ in the self-attention module play different roles in the generation process. $K$ and $V$ are more directly involved in appearance feature matching and aggregation, whereas $Q$ affects the query relationships and spatial organization among tokens \cite{chung2024style,zhou2025attention}. Therefore, we only suppress the target style responses captured by $U_{style}$ along the $K$ and $V$ paths in the self-attention module. Specifically, we first compute the projection of any token feature $x_i\in\mathbb{R}^{D}$ onto the corresponding subspace:
\begin{equation}
	\small
\operatorname{proj}_{U_{style}}(x_i) =  (x_i U_{style})U_{style}^\top
\end{equation}
Subsequently, we attenuate the projected style related response by subtracting its projection from the original $K$ and $V$ matrices, resulting in the edited matrices $K'$ and $V'$:
\begin{equation}
	\small
	K_i' = K_i - \gamma_i \cdot \operatorname{proj}_{U_{\mathrm{style}}}(K_i)
\end{equation}
\begin{equation}
	\small
	V_i' = V_i - \gamma_i \cdot \operatorname{proj}_{U_{\mathrm{style}}}(V_i).
\end{equation}
By applying suppression only to the $K$ and $V$ matrices, we suppress style related features such as texture and brushstrokes while reducing unnecessary changes to the content structure. The edited representations are fed back into the original self-attention computation:
\begin{equation}
	\small
	\text{Attention}(Q, K', V') = \operatorname{softmax}\left(\frac{QK'^\top}{\sqrt{d_h}}\right)V'
\end{equation}
We illustrate the complete algorithmic framework in Figure \ref{figure3}.

\section{Experiments}
\label{sec: experiments}

\subsection{Implementation Details}

In this section, we conduct a comprehensive evaluation of the proposed CAPE framework against several baseline methods, including SPEED \cite{li2026speed}, ESD-u \cite{gandikota2023erasing}, ESD-x \cite{gandikota2023erasing}, FMN \cite{zhang2024forget}, MACE \cite{lu2024mace}, P2P \cite{hertz2022prompt}, and SELECT \cite{zhang2025beyond}. All main comparison and ablation experiments use SD v1.4 \cite{rombach2022high}, and images are generated at $512\times512$ resolution with 50 DDIM denoising steps \cite{song2020denoising}.  The experiments were conducted on NVIDIA RTX A6000. 

Hyperparameter Settings. For style subspace estimation, we extract intermediate features from the shallow U-Net block (down\_blocks[0]) at feature-probe index $s=100$ on the 1000 step DDIM scheduler. We set the content penalty weight to $\lambda=0.2$ and retain the top $r=15$ generalized eigen-directions to construct $U_{\mathrm{style}}$. For the Adaptive Suppression Controller, we use $\alpha_{\min}=0.6$, $\alpha_{\max}=1.8$, $k=40$, and $\tau=0.3$, with response weights $(w_q,w_k,w_v)=(0.20,0.35,0.45)$.

\subsection{Evaluation Metrics}

\subsubsection{CLIP Score} 
We adopt CLIP \cite{radford2021learning} to separately evaluate the target artist style response and content semantic preservation. To reduce the interference between content and style within the same prompt, we construct independent text templates for target style and content, respectively. Style prompts only contain descriptions of the target artist style, while content prompts only describe the target content. The CLIP cosine similarity between generated images and the corresponding text templates are denoted as $cs_{style}$ and $cs_{content}$, respectively. Furthermore, we use Style Suppression $S_{sup}$ to represent the reduction of the target artist style score before and after generation:
\begin{equation}
	\small
	S_{sup}=cs_{style}^{original}-cs_{style}^{edited}
\end{equation}
A larger $S_{sup}$ indicates a greater decrease in the target artist style response, while a higher $cs_{content}$ indicates better consistency between the generated content and the requested semantics.

\subsubsection{Differential-LPIPS}
Learned Perceptual Image Patch Similarity (LPIPS) \cite{zhang2018unreasonable} is used to measure the difference between two images in the perceptual feature space. A larger value indicates a greater perceptual difference between the two images. Traditional LPIPS can only provide the overall perceptual difference between two images, without distinguishing whether this difference originates from style suppression or content degradation. We proposed the Differential-LPIPS (DLPIPS). By calculating the differences between each pair of samples to measure the perceived differences, we evaluation of style suppression performance is conducted from three dimensions:

\textbf{Style Generalization}. Generate the reference image $I_{\text{ref}}^{\text{gene}}$ using a variant of the original prompt. Calculate the LPIPS score $L_{gene}$ between the edited image $I_{edit}$ and $I_{\text{ref}}^{\text{gene}}$. When $L_{gene}$ is larger, it indicates that although there are slight changes in the expression, the scheme can still generalize the suppression effect to a fixed artist style.

\textbf{Style Suppression Effectiveness}. Using prompts of different content but with the same style to generate reference image $I_{\text{ref}}^{\text{style}}$, calculate the LPIPS ($L_{\text{base}}^{\text{style}}$) of original image $I_{\text{ori}}$ and $I_{\text{ref}}^{\text{style}}$ as the baseline, which measures the perceptual difference caused by the different content. we calculate the LPIPS ($L_{\text{edit}}^{\text{style}}$) of $I_{edit}$ and $I_{\text{ref}}^{\text{style}}$, obtaining $C_{style}$:
\begin{equation}
	\small
	C_{style}= L_{edit}^{style}-L_{base}^{style}
\end{equation}
This difference measures the additional perceptual distance caused by suppressing the shared target style. A larger $C_{style}$ indicates a stronger perceptual deviation from the shared target style reference after accounting for the content induced baseline.

\textbf{Content Preservation}. Content Preservation focuses on the degree of content preservation during the suppression process. We generate a reference image $I_{\text{ref}}^{\text{cont}}$ and calculate its LPIPS score ($L_{\text{base}}^{\text{cont}}$) with the original image $I_{\text{ori}}$ as a baseline. This baseline measures the perceptual difference caused by the different styles.
Subsequently, we calculate the LPIPS score ($L_{\text{edit}}^{\text{cont}}$) between the reference image $I_{\text{ref}}^{\text{cont}}$ and the erased image $I_{\text{edit}}$. To better measure the additional content degradation introduced during the suppression process, we calculate the difference:
\begin{equation}
	\small
	C_{content}= L_{edit}^{cont}-L_{base}^{cont}
\end{equation}
This difference measures the additional perceptual difference caused by content loss during style suppression. To make the metric more stable and robust, we use multiple reference styles to average the baseline.

\subsubsection{CSD Style Feature}
Contrastive Style Descriptors(CSD) \cite{somepalli2024measuring} provide a style representation that is relatively robust to content variations. We calculate the cosine distance before and after suppression to measure the artistic style similarity change between generated images:
\begin{equation}
	\small
	 C_{\mathrm{CSD}} = 1- \cos \left( f_{\mathrm{CSD}} \left( I_{\mathrm{orig}} \right), f_{\mathrm{CSD}} \left( I_{\mathrm{edit}} \right) \right) 
\end{equation}
A larger $C_{\mathrm{CSD}}$ indicates a more significant change in the CSD style representation before and after editing. It should be interpreted together with content preservation metrics, since global image reconstruction may also increase this distance.

\subsubsection{VGG-19 Gram Style Distance}
We adopt the multi-layer Gram representation of the pretrained VGG-19 \cite{simonyan2014very} to measure changes in color, texture, and local brushstroke statistics. This representation was originally introduced to characterize correlations between convolutional feature channels and has been widely used for artistic style modeling \cite{gatys2016image}. We calculate the Gram distance between two images at layer $\ell$:
\begin{equation}
	\small
 C_{\mathrm{VGG}} = \frac{1}{|\mathcal{L}|} \sum_{\ell\in\mathcal{L}} \left\| G_{\ell}\left(I_{ori}\right) - G_{\ell}\left(I_{edit}\right) \right\|_{F}^{2}
\end{equation}
where $G_{\ell}(\cdot)$ denotes the Gram matrix of VGG features at layer $\ell$. A larger $C_{\mathrm{VGG}}$ indicates more significant changes in multi-layer texture statistics before and after editing.

\begin{figure*}[h] 
	\centering 
	\includegraphics[width=1\textwidth]{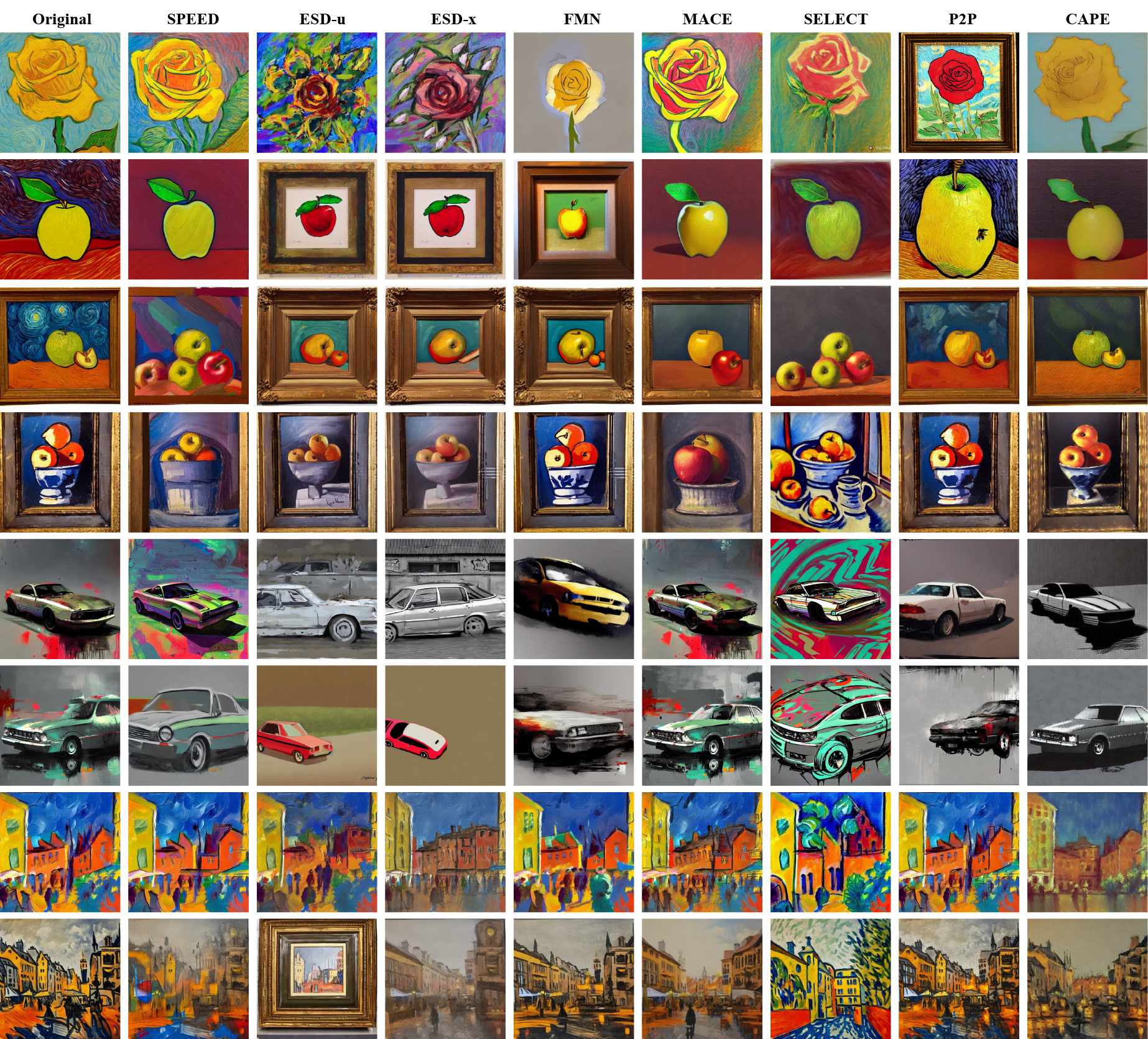} 
	\caption{Qualitative comparison with representative baselines on Van Gogh, Maurice De Vlaminck, Adrian Ghenie, and Charles Camoin. CAPE suppresses characteristic style patterns while preserving the requested content and scene structure more consistently across diverse examples.} 
	\label{figure4} 
\end{figure*}

\subsubsection{CLIP Object Accuracy}
To directly evaluate whether the main semantic entities remain recognizable after editing, we adopt an object recognition metric based on the CLIP model. For the entity set $\mathcal E(p)$ contained in prompt $p$, an entity $e$ is considered successfully preserved if it appears in the CLIP recognition list of the generated image. A higher CLIP Accuracy indicates that more semantic entities remain correctly recognized after editing, reflecting the preservation degree of content entities. We report the fraction of annotated prompt entities whose label or predefined semantic alias appears in the CLIP Top-8 predictions, averaged over all evaluated prompts.

\subsubsection{Content Consistency}
Since content preservation involves multiple aspects, including semantic entities, spatial layouts, and low-level structures, a single metric cannot fully characterize it. Therefore, we additionally report DINO Content Consistency and SSIM \cite{wang2004ssim} as supplementary references. DINOv2 \cite{oquab2023dinov2} is used to measure the consistency of high-level visual representations between original and edited images. A higher $C_{DINO}$ indicates better feature consistency before and after editing. Structural Similarity Index Measure (SSIM) evaluates the consistency between paired images from three aspects, including luminance, contrast, and local structure. A higher SSIM indicates better preservation of the original image structure. We use these two metrics as supplementary indicators for content preservation.

\subsection{Standard Performance Evaluation}
We evaluate CAPE on seven artist styles with distinct visual characteristics. For each style, we use three prompts with different content and five random seeds. The quantitative evaluation focuses on two aspects: target style suppression and content preservation. For target style suppression, we use VGG19 Gram style distance $C_{VGG}$, DLPIPS $C_{\mathrm{style}}$, CSD style distance $C_{CSD}$, and CLIP Score $S_{sup}$ to quantify the suppression of the target style. As shown in Table \ref{tab3}, CAPE achieves the best performance on VGG19 distance and DLPIPS $C_{\mathrm{style}}$, which mainly reflect changes in local texture, brushstrokes, color organization, and surface statistics. In comparison, MACE obtains the largest CLIP score drop $S_{sup}$, while ESD-x achieves the largest $C_{CSD}$. These metrics are more sensitive to changes in the overall visual response of the target artist concept, and therefore larger global appearance changes, as shown in Figure \ref{figure4}, may also lead to higher score variations.

For content preservation, we use the CLIP Accuracy metric to evaluate whether the main content entities in prompts remain recognizable after editing. This metric measures the proportion of entities contained in the prompt that can still be correctly recognized after editing. CAPE maintains strong performance in content recognition, indicating that it can preserve semantic entities while producing more noticeable suppression of local rendering characteristics associated with the target style.

Since object recognition metrics can only verify whether the main subjects remain present and cannot reliably evaluate other visual elements such as background layouts and relative spatial relationships among entities, we further conduct qualitative evaluations with multiple examples. The results are shown in Figure \ref{figure4}. CAPE effectively weakens the brushstrokes, textures, and local rendering patterns associated with the target artist style, while preserving the major visual elements unrelated to the target style in the original image. The category of the main subject and the primary objects remain stable, and the essential scene composition in the background is preserved. In comparison, although some baselines can substantially reduce the target style response, they may introduce noticeable changes in subject appearance, background structure, or scene layout. Other methods may better preserve the original structure in some cases, but still retain strong target style characteristics.

\begin{table}[t]
	\centering
	\caption{Standard performance evaluation results. CAPE effectively suppresses target style characteristics while preserving semantic content more consistently than the compared methods.}
	\label{tab3}
	
	\footnotesize
	\setlength{\tabcolsep}{3pt}
	\renewcommand{\arraystretch}{0.9}
	
	\begin{tabular*}{\columnwidth}{@{\extracolsep{\fill}}lccccc}
		\toprule
		
		Method
		&
		$C_{VGG}\uparrow$
		&
		$C_{\mathrm{style}}\uparrow$
		&
		$C_{CSD}\uparrow$
		&
		$S_{sup}\uparrow$
		&
		CLIP Acc$\uparrow$
		\\
		
		\midrule

		SPEED \cite{li2026speed}
		&0.0566
		&0.0485
		&0.1368
		&2.55
		&0.9587
		\\
		
		ESD-u \cite{gandikota2023erasing}
		&0.1148
		&0.0747
		&0.2346
		&2.67
		&0.9365
		\\
		
		ESD-x \cite{gandikota2023erasing}
		&0.0971
		&0.1005
		&\textbf{0.3657}
		&4.30
		&0.9341
		\\
		
		FMN \cite{zhang2024forget}
		&0.0871
		&0.0653
		&0.1790
		&2.65
		&0.9675
		\\
		
		MACE \cite{lu2024mace}
		&0.1077
		&0.0907
		&0.2744
		&\textbf{4.88}
		&0.9730
		\\
		
		P2P \cite{hertz2022prompt}
		&0.0115
		&0.0195
		&0.0596
		&0.66
		&0.9000
		\\
		 
		SELECT \cite{zhang2025beyond}
		&0.0545
		&0.0474
		&0.1586
		&2.94
		&0.9349
		\\
		
		Ours
		& \textbf{0.1489}
		& \textbf{0.1524}
		&0.2478
		&3.01
		&\textbf{0.9754}
		\\
		\bottomrule
		
	\end{tabular*}
	
\end{table}

Combining the quantitative results and qualitative comparisons, CAPE primarily affects the rendering characteristics of the target artist style. Such style suppression does not require substantial modifications to the subject, background, or geometric structures, thereby achieving the goal of artist style purification.

\begin{figure}[h] 
	\centering 
	\includegraphics[width=1\columnwidth]{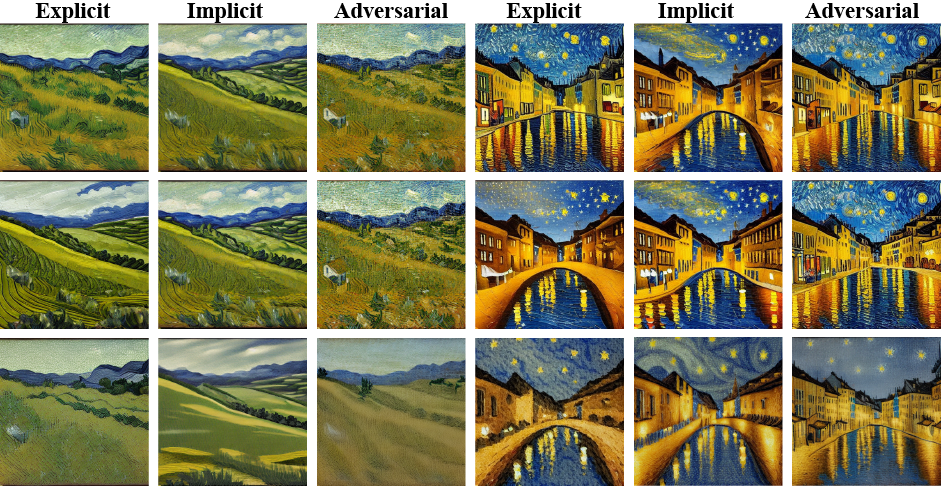} 
	\caption{Qualitative comparison with prompt filtering under explicit, implicit, and adversarial requests for Van Gogh style. From top to bottom, they are: original, Filtered, CAPE. For each content scene, the three prompt variants target the same artist style while using different textual expressions. Prompt filtering fails under implicit or obfuscated artist cues, whereas CAPE suppresses target-style patterns across all three prompt forms.} 
	\label{figure5} 
\end{figure}

We introduce an additional baseline, prompt filtering, which removes artist style related expressions from the input text before generation. This baseline is included to directly investigate whether simply removing style mentions from prompts can reliably prevent artist style mimicry. We evaluate three types of prompts: explicit prompts, implicit prompts, and adversarial prompts. Explicit prompts directly include the artist name. Implicit prompts do not explicitly mention the artist name but describe style characteristics through visual cues such as brushstrokes and color patterns. Adversarial prompts obfuscate artist names through spelling variations, multilingual expressions, or Unicode perturbations. The experimental results are shown in Figure \ref{figure5}. Prompt filtering fails to remove the target style under implicit and adversarial prompts, whereas CAPE suppresses the same target style patterns across all three prompt forms.

\subsection{Ablations}
\subsubsection{Triplet Estimation}

We construct triplets to compute the dominant style direction set $U_{\mathrm{style}}$, and reduce the components supported by shared content relations in the AP direction through the AN relation. To verify whether the $U_{\mathrm{style}}$ estimated by triplets truly captures generalizable target artist style information rather than being only applicable to specific prompts, specific contents, or random directions, we design five ablation settings: Full Triplet, AP only ($\lambda=0$), AP/AN Swap (swapping Positive and Negative), Random (randomly generating an orthogonal low rank basis with the same dimension as $U_{\mathrm{style}}$), and Shuffled Positive (randomly permuting the rows of $P$). Using the $U_{\mathrm{style}}$ obtained from different strategies, we calculate new triplets with unseen positive contents, negative styles, and seeds:
\begin{equation}
	\small
	R_{AP}^{h}=\Sigma_{AP}^{h}\Sigma_{PA}^{h}
	\qquad
	R_{AN}^{h}=\Sigma_{AN}^{h}\Sigma_{NA}^{h}
	\qquad
	B_h=\Sigma_{AA}^{h}+\epsilon I I
\end{equation}
We compute the following metrics for the $U_{\mathrm{style}}$ obtained by each strategy:
\begin{equation}
	\small
H_{\mathrm{style}}(U)
	=
	\frac{1}{r}
	\operatorname{Tr}
	\left[
	\left(U^{T}B_hU\right)^{-1}
	U^{T}R_{AP}^{h}U
	\right]
\end{equation}
\begin{equation}
	\small
H_{\mathrm{content}}(U)
	=
	\frac{1}{r}
	\operatorname{Tr}
	\left[
	\left(U^{T}B_hU\right)^{-1}
	U^{T}R_{AN}^{h}U
	\right]
\end{equation}
$H_{\mathrm{style}}$ and $H_{\mathrm{content}}$ measure the extent to which these directions capture the target style and content on unseen samples, respectively. Larger values indicate stronger responses of the corresponding relation in the estimated directions. The complete results are shown in Table \ref{tab4}. First, Random and Shuffled Positive exhibit substantially weaker responses and suppression performance than the other strategies, indicating that effective style control directions do not originate from arbitrary low dimensional subspaces or random patch correspondences, but depend on the structured relations provided by the triplets.

\begin{table}[t]
	\centering
	\caption{Comparison of different triplet estimation strategies for target-style suppression and content preservation.}
	\label{tab4}
	\footnotesize
	\setlength{\tabcolsep}{3.5pt}
	
	\begin{tabular}{lccccc}
		\toprule
		
		\textbf{Estimation}
		&
		$H_{\mathrm{style}}\uparrow$
		&
		$H_{\mathrm{content}}\downarrow$
		&
		$S_{sup}\uparrow$
		&
		$cs_{\mathrm{style}}\downarrow$
		&
		$cs_{\mathrm{content}}\uparrow$
		\\
		
		\midrule

		Random
		&
		6.643
		&
		6.654
		&
		1.7
		&
		24.29
		&
		30.83
		\\
		
		Shuffled
		&
		0.955
		&
		0.956
		&
		1.9
		&
		24.08
		&
		30.59
		\\
		\midrule
		AP-only
		&
		\textbf{10.164}
		&
		10.181
		&
		2.6
		&
		23.38
		&
		\textbf{31.28}
		\\

		Swap
		&
		10.152
		&
		10.174
		&
		2.6
		&
		23.41
		&
		31.12
		
		\\
		Triplet
		&
		10.157
		&
		\textbf{10.170}
		&
		\textbf{2.7}
		&
		\textbf{23.36}
		&
		31.19
		\\
		
		\bottomrule
	\end{tabular}
\end{table}

Second, AP only achieves the highest $H_{\mathrm{style}}$ while also obtaining a higher $H_{\mathrm{content}}$, indicating that relying only on AP can preserve stronger target style information but lacks constraints on content related directions. In comparison, Full Triplet achieves an $H_{\mathrm{style}}$ close to AP only while obtaining the lowest $H_{\mathrm{content}}$. It is worth noting that the differences between the two settings in $H_{\mathrm{content}}$ and $H_{\mathrm{style}}$ are relatively small, which is consistent with the strong shared relationship between artist style and content in generation features. Artist style is not only expressed through appearance attributes such as texture, brushstrokes, and color, but also jointly affects local object contours, surface organization, and spatial structures. For example, apples generated with Van Gogh and Monet styles may exhibit different structural appearances rather than completely following the visual representation of real world objects. Therefore, the directions supported by AP and AN relations are not completely independent. The AN based penalty term reduces directions jointly supported by content relations while preserving the main style response. As a result, Full Triplet maintains a similar $H_{\mathrm{style}}$ to AP only, while further reducing $H_{\mathrm{content}}$, achieving the best suppression performance and maintaining a content score close to AP only, thereby obtaining a better suppression and preservation trade off. In addition, across multiple concepts and parameter settings, Full Triplet consistently outperforms Swap with relatively small but stable margins across multiple metrics, further reflecting that style and content share substantial structures in the current feature space.

\begin{figure}[h]
	\centering
	\includegraphics[width=1\columnwidth]{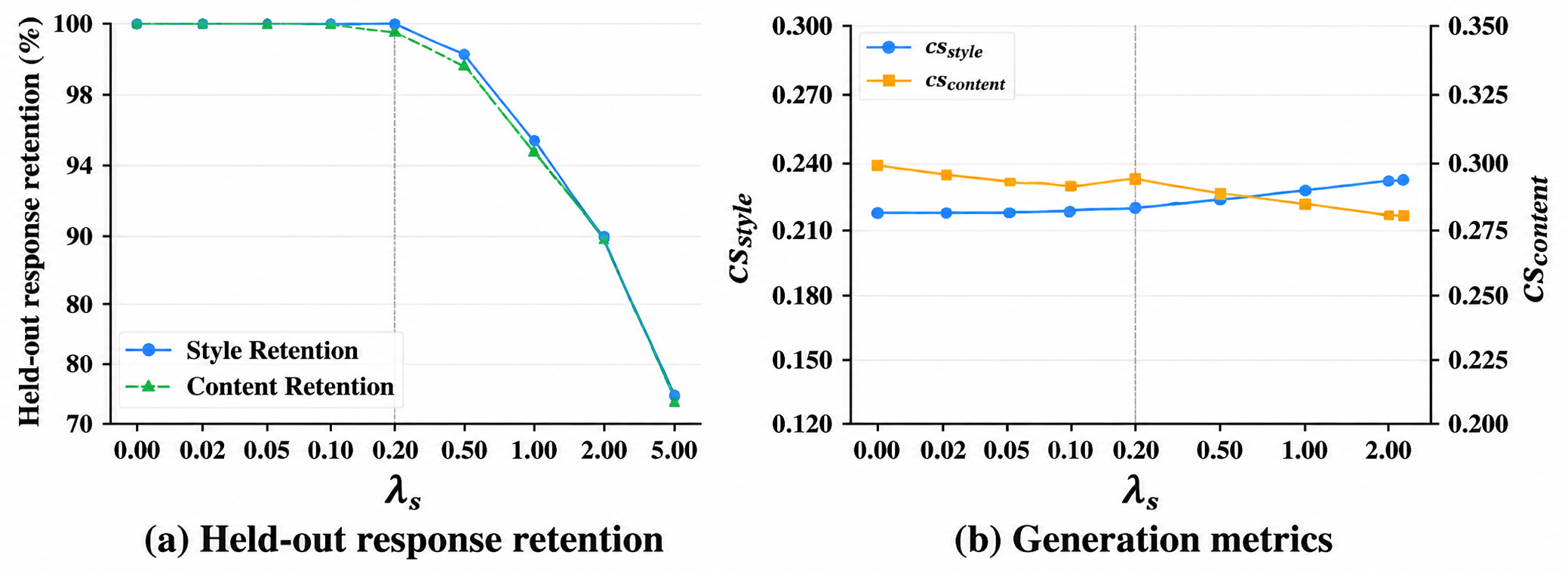} 
	\caption{Sensitivity analysis of $\lambda$. (a) Style and content response retention. (b) Target-style and content scores under different $\lambda$s.}
	\label{figure6}
\end{figure}

We further investigate the sensitivity of the triplet penalty coefficient $\lambda$. The results are shown in Figure \ref{figure6}. As the content penalty coefficient $\lambda$ increases, $H_{content}$ and $H_{style}$ remain close, while $H_{content}$ decreases faster than $H_{style}$, indicating that the AN term effectively reduces content related responses. In addition, within the range of $\lambda \in [0.05,0.2]$, style and content responses as well as suppression performance remain stable and close, indicating that CAPE performance does not rely on fine tuning of $\lambda$.

\subsubsection{Adaptive Suppression Controller}

\begin{figure}[h]
	\centering
	\includegraphics[width=1\columnwidth]{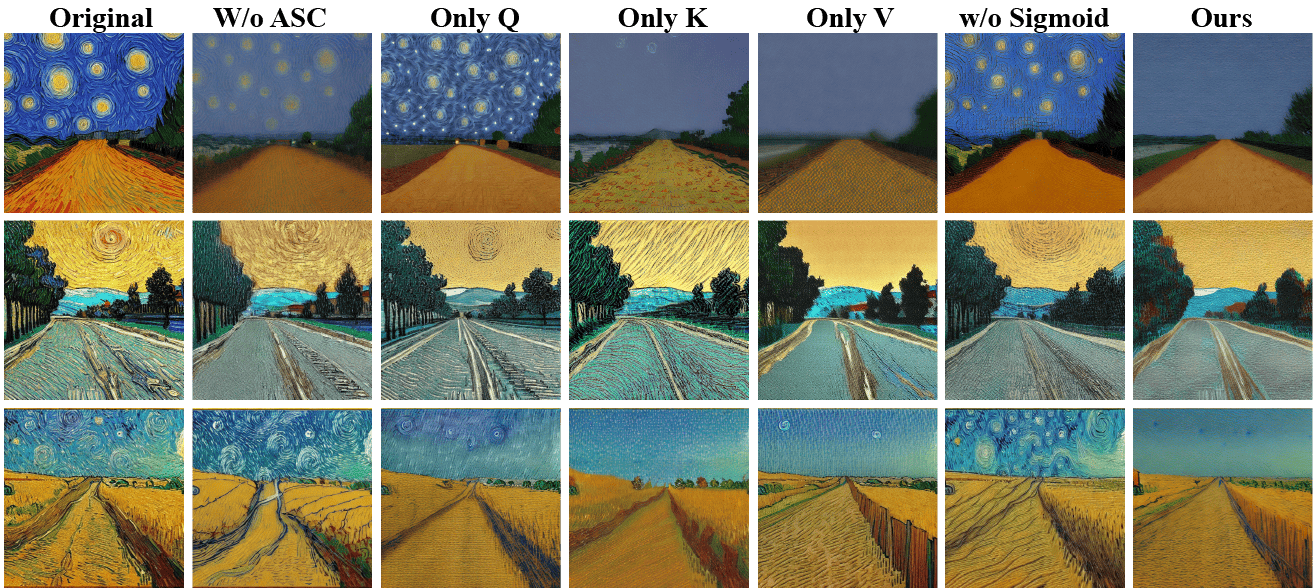} 
	\caption{Qualitative comparison of Adaptive Suppression Controller variants. The ASC provides more consistent suppression across spatially heterogeneous style regions while retaining the principal content structure.}
	\label{figure7}
\end{figure}

The Adaptive Suppression Controller aggregates style scores from Q, K, and V with weighted summation and employs a Sigmoid function to achieve adaptive control across different regions. We conduct ablation studies from two perspectives, including style score composition and adaptive control. Specifically, we compare using only Q, K, or V for style score computation, removing the Sigmoid mapping and replacing it with linear mapping, and completely removing ASC. The results are shown in Figure \ref{figure7} and Table \ref{tab5}. The complete adaptive suppression controller achieves a better overall balance between target style suppression and content preservation. Without adaptive control, different regions may suffer from residual style patterns or content degradation.

\begin{table}[t]
	\centering
	\caption{Comparison of different Adaptive Suppression Controller variants. w/o Sigmoid means removing the Sigmoid mapping and replacing it with linear mapping.}
	\label{tab5}
	\footnotesize
	\setlength{\tabcolsep}{3.5pt}
	
	\begin{tabular}{lccccc}
		\toprule
		
		Metric
		&
		$L_{gene}\uparrow$
		&
		$C_{style}\uparrow$
		&
		$C_{content}\downarrow$
		&
		$cs_{style}\downarrow$
		&
		$cs_{content}\uparrow$
		\\
		
		\midrule
		
		only\_K
		&
		0.631
		&
		0.200
		&
		-0.027
		&
		29.52
		&
		26.41
		\\
		
		only\_V
		&
		0.649
		&
		0.239
		&
		-0.013
		&
		29.31
		&
		26.91
		\\
		
		only\_Q
		&
		0.592
		&
		0.177
		&
		-0.012
		&
		29.73
		&
		26.22
		\\
		
		w/o Sigmoid
		&
		0.591
		&
		0.154
		&
		0.001
		&
		30.40
		&
		26.69
		\\
		
		Ours
		&
		\textbf{0.664}
		&
		\textbf{0.251}
		&
		\textbf{-0.043}
		&
		\textbf{28.79}
		&
		\textbf{27.24}
		\\
		
		\bottomrule
	\end{tabular}
\end{table}

Furthermore, we verify whether the patch level style concentration estimated by ASC can effectively localize the regions requiring stronger suppression. Based on the patch style concentration, we select the top 20\% patches with the highest concentration ($Top$), the bottom 20\% patches with the lowest concentration ($Bottom$), and randomly selected 20\% patches ($random$), and apply KV suppression with the same strength to these patches. The results are shown in Table \ref{tab6}. Patches with higher style concentration achieve the lowest residual style score and the highest content score, indicating that the style concentration estimated by ASC can effectively identify regions more relevant for style suppression rather than selecting patches arbitrarily.

\begin{table}[t]
	\centering
	\caption{Effect of patch selection based on style concentration.}
	\label{tab6}
	
	\footnotesize
	\setlength{\tabcolsep}{5pt}
	\renewcommand{\arraystretch}{0.9}
	
	\begin{tabular*}{\columnwidth}{@{\extracolsep{\fill}}lcc}
		\toprule
		
		\textbf{Selected patches}
		&
		$cs_{\mathrm{style}}\downarrow$
		&
		$cs_{\mathrm{content}}\uparrow$
		\\
		
		\midrule
		
		Top 20\%
		&
		\textbf{28.78}
		&
		\textbf{29.76}
		\\
		
		Random 20\%
		&
		29.13
		&
		29.56
		\\
		
		Bottom 20\%
		&
		29.29
		&
		29.34
		\\
		
		\bottomrule
		
	\end{tabular*}
	
\end{table}

\subsubsection{Attention KV Intervention}

\begin{figure}[h]
	\centering
	\includegraphics[width=1\columnwidth]{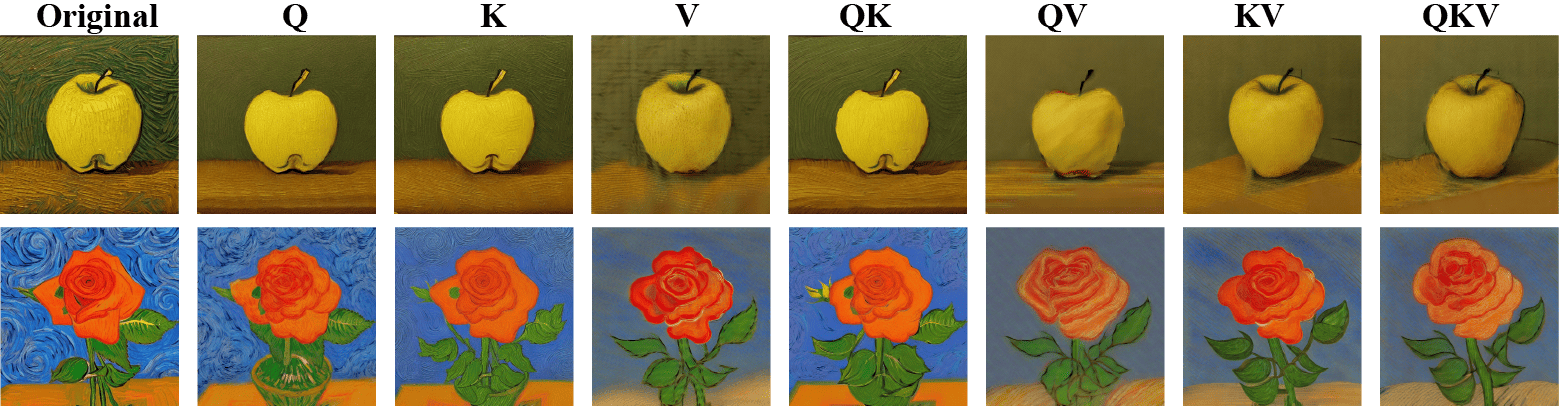} 
	\caption{Qualitative comparison of attention intervention pathways. KV achieves target-style suppression visually comparable to QKV without directly intervening on Q.}
	\label{figure8}
\end{figure}

CAPE performs style suppression only on the K and V branches. To validate this design choice, we compare seven attention intervention pathways: Q, K, V, QK, QV, KV, and QKV. We use $L_{gene}$, $C_{style}$, and $cs_{style}$ to evaluate style suppression performance, and use $C_{content}$ and $cs_{content}$ to measure content changes and semantic preservation. We further report $C_{DINO}$ and SSIM to evaluate high level visual semantics and structural consistency. The complete quantitative results are shown in Table \ref{tab7}. The results show that independently intervening on Q or K provides relatively limited target style suppression, while combinations involving V achieve more significant style suppression. QKV achieves the strongest overall suppression, but also introduces the largest perceptual content change and exhibits more obvious structural degradation in DINO and SSIM. In comparison, the KV strategy achieves suppression performance comparable to or better than QKV across multiple style suppression and content preservation metrics, while obtaining better structural preservation scores than QKV. Qualitative results in Figure \ref{figure8} further show that individual branch interventions and other combinations still retain partial target style brushstrokes. KV achieves a visual suppression effect close to QKV, whereas additionally intervening on Q provides limited extra suppression gain while introducing more noticeable structural changes.

\begin{table*}[t]
	\centering
	\caption{Comparison of attention intervention pathways. Best metric values are shown in bold, and second-best metric values are underlined.}
	\label{tab7}
	
	\footnotesize
	\setlength{\tabcolsep}{4pt}
	\renewcommand{\arraystretch}{0.9}
	
	\begin{tabular*}{\textwidth}{@{\extracolsep{\fill}}lccccccc}
		\toprule
		
		Pathway
		&
		$L_{\mathrm{gene}}\uparrow$
		&
		$C_{\mathrm{style}}\uparrow$
		&
		$cs_{\mathrm{style}}\downarrow$
		&
		$C_{\mathrm{content}}\downarrow$
		&
		$cs_{\mathrm{content}}\uparrow$
		&
		$C_{\mathrm{DINO}}\uparrow$
		&
		SSIM$\uparrow$
		\\
		
		\midrule
		
		Q
		&0.3709
		&0.0600
		&32.42
		&\underline{0.0244}
		&\textbf{32.95}
		&\textbf{0.8402}
		&\textbf{0.6425}
		\\
		
		K
		&0.3792
		&0.0553
		&32.47
		&\textbf{0.0203}
		&\underline{32.61}
		&0.8331
		&\underline{0.6164}
		\\
		
		V
		&0.4426
		&0.1059
		&31.33
		&0.0565
		&32.24
		&0.7408
		&0.5113
		\\
		
		QK
		&0.4080
		&0.0759
		&31.93
		&0.0347
		&32.45
		&\underline{0.8336}
		&0.5844
		\\
		
		QV
		&0.5153
		&0.1473
		&31.56
		&0.0893
		&32.30
		&0.7145
		&0.4743
		\\
		
		\textbf{KV}
		&\underline{0.5280}
		&\underline{0.1581}
		&\underline{30.98}
		&0.0885
		&31.86
		&0.7194
		&0.4707
		\\
		
		QKV
		&\textbf{0.5526}
		&\textbf{0.1713}
		&\textbf{30.71}
		&0.1056
		&32.43
		&0.6902
		&0.4505
		\\
		
		\bottomrule
		
	\end{tabular*}
	
\end{table*}

Furthermore, to eliminate the influence of different suppression strengths among pathways, we adjust the intervention strength of each pathway to achieve the same normalized $C_{style}$ level, and then compare their content and structural preservation performance. As shown in Table \ref{tab8}, KV achieves the best or second best performance across multiple metrics measuring semantic, contour, and perceptual content perturbations. Although QKV provides stronger suppression capability, it does not achieve more stable structural preservation.
\begin{table}[t]
	\centering
	\caption{Content preservation results of attention pathways. Best metric values are shown in bold, and second-best metric values are underlined.}
	\label{tab8}
	
	\footnotesize
	\setlength{\tabcolsep}{3pt}
	\renewcommand{\arraystretch}{0.9}
	
	\begin{tabular*}{\columnwidth}{@{\extracolsep{\fill}}lcccc}
		\toprule
		
		Pathway
		&
		$C_{\mathrm{content}}\downarrow$
		&
		$cs_{\mathrm{content}}\uparrow$
		&
		$C_{\mathrm{DINO}}\uparrow$
		&
		SSIM$\uparrow$
		\\
		
		\midrule
		
		Q
		&0.0620
		&32.14
		&0.7477
		&\textbf{0.5520}
		\\
		
		K
		&0.0616
		&31.83
		&0.7388
		&0.5059
		\\
		
		V
		&\underline{0.0560}
		&32.30
		&0.7492
		&0.5122
		\\
		
		QK
		&0.0664
		&30.83
		&0.6771
		&0.5021
		\\
		
		QV
		&0.0707
		&32.26
		&0.7641
		&0.5245
		\\
		
		\textbf{KV}
		&\textbf{0.0520}
		&\underline{32.36}
		&\textbf{0.7890}
		&\underline{0.5398}
		\\
		
		QKV
		&0.0637
		&\textbf{32.79}
		&\underline{0.7851}
		&0.5326
		\\
		
		\bottomrule
		
	\end{tabular*}
	
\end{table}

\subsubsection{Cross-architecture Transfer}

To evaluate the cross architecture adaptability of CAPE, we further transfer it from the U Net based Stable Diffusion v1.4 to the Transformer based FLUX \cite{esser2024scaling}. Considering the trajectory based generation mechanism of FLUX based on flow matching and MMDiT, we compare the differences produced by triplet conditions along the generation trajectory under the shared current state, and adapt the style suppression during inference to the image token generation pathway of FLUX.

\begin{figure}[h]
	\centering
	\includegraphics[width=1\columnwidth]{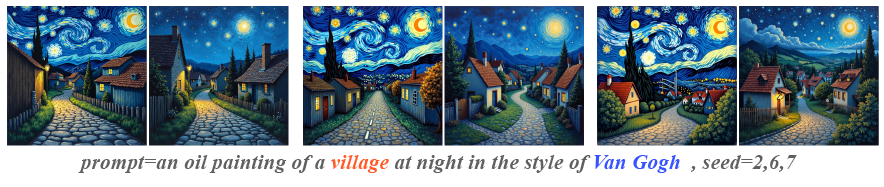} 
	\caption{Qualitative cross architecture evaluation on FLUX. CAPE suppresses the characteristic swirling brush strokes and high saturation starry sky textures while preserving the village layout, object structure, and global scene composition.}
	\label{figure9}
\end{figure}

As shown in Figure \ref{figure9}, CAPE can suppress Van Gogh style related rendering patterns, including swirling brushstrokes, texture structures, and color organization, while preserving the main scene semantic structure, validating the effectiveness of transferring CAPE to Transformer based diffusion architectures.

\subsubsection{Computational Efficiency}

We further compare the one time offline processing cost and single image inference cost of different methods. Offline End-to-End(E2E) time covers the complete process required before each method can be applied, including data preparation, training, fine tuning, weight editing, or feature direction estimation. Inference time records the single image denoising generation latency. The incremental peak GPU memory $\Delta$VRAM is calculated by subtracting the device memory baseline before execution from the peak memory usage during execution. The experimental results are shown in Table \ref{tab9}.

Among methods requiring offline preparation, CAPE achieves the second lowest E2E time, only slower than SPEED. Meanwhile, CAPE obtains the lowest offline incremental peak memory consumption, reducing the memory overhead by approximately 70.0\% compared with other training based or weight editing methods. During inference, the latency and memory consumption of CAPE remain close to those of the original SD1.4 model, indicating that the ASC response computation and K/V suppression introduce no significant additional overhead. Compared with P2P, which is also an inference time intervention method, CAPE reduces the single image latency by approximately 59.6\% and the incremental peak memory consumption by approximately 41.8\%.

\begin{table*}[t] 
	\centering 
	\caption{Computational efficiency measured for one target concept, one prompt, and one seed.} 
	\label{tab9} 
	
	\footnotesize 
	\setlength{\tabcolsep}{8pt} 
	
	\begin{tabular*}{\textwidth}{@{\extracolsep{\fill}}lcccc} 
		\toprule 
		
		Method 
		& 
		Offline E2E (s)$\downarrow$ 
		& 
		Offline $\Delta$VRAM (MiB)$\downarrow$ 
		& 
		Inference (s/img)$\downarrow$ 
		& 
		Inference $\Delta$VRAM (MiB)$\downarrow$ 
		\\ 
		
		\midrule 
		
		SD1.4 
		&-- 
		&-- 
		&2.268 
		&\textbf{3272} 
		\\ 
		
		SPEED \cite{li2026speed}
		&\textbf{56.647} 
		&17116 
		&2.164 
		&\textbf{3272} 
		\\ 
		
		ESD-u \cite{gandikota2023erasing}
		&1217.075 
		&15423 
		&2.131 
		&\textbf{3272} 
		\\ 
		
		ESD-x \cite{gandikota2023erasing}
		&897.789 
		&10320 
		&2.130 
		&\textbf{3272} 
		\\ 
		
		FMN \cite{zhang2024forget}
		&262.485 
		&18838 
		&2.145 
		&\textbf{3272} 
		\\ 
		
		MACE \cite{lu2024mace}
		&180.005 
		&17844 
		&2.238 
		&\textbf{3272} 
		\\ 
		
		SELECT \cite{zhang2025beyond}
		&97.729 
		&15376 
		&\textbf{2.129} 
		&\textbf{3272} 
		\\ 
		
		P2P \cite{hertz2022prompt}
		&-- 
		&-- 
		&5.488 
		&5626 
		\\ 
		
		Ours 
		&81.306 
		&\textbf{3098} 
		&2.219 
		&3274 
		\\ 
		
		\bottomrule 
		
	\end{tabular*} 
\end{table*}

\subsection{Sensitivity Analysis}

To verify the impact of key hyperparameters on the robustness of our method, we conduct sensitivity tests on three groups of hyperparameters while keeping all other settings unchanged, including the eigenbasis rank $r$, the feature extraction diffusion position $s$, and the triplet reference choices for positive content and negative style.

\subsubsection{Style Subspace Dimension}

\begin{figure}[h]
	\centering
	\includegraphics[width=1\columnwidth]{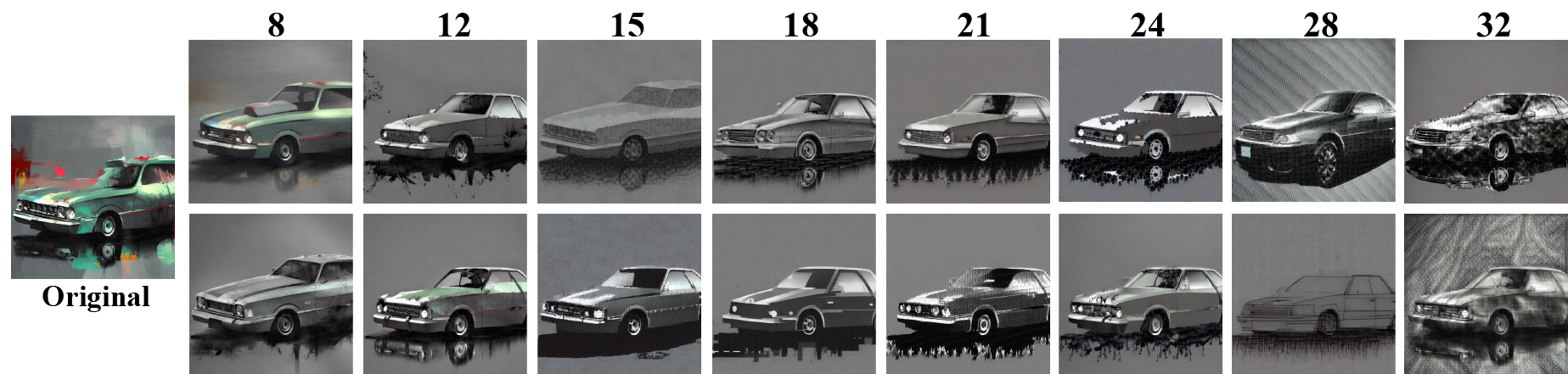} 
	\caption{Style subspace dimension sensitivity. Small $r$ retains artist specific color and brush traits, large $r$ introduces high frequency artifacts. The mid-range [12,21] consistently achieves clean style removal with high content fidelity.}
	\label{figure10}
\end{figure}

We construct $U_{style}$ using the eigenvectors corresponding to the first $r$ largest eigenvalues. To verify whether directions with larger generalized eigenvalues indeed contain more effective information for suppressing target styles, we first select five artist styles and evaluate eight settings. Specifically, we select the first $\{8,12,15,18,21,24,28,32\}$ eigenvectors ranked by generalized eigenvalues to construct the style subspace. The results are shown in Figure \ref{figure10}. When the dimension used to capture the style subspace is too small, although prominent brushstrokes and exaggerated color patterns can be suppressed, residual style textures remain in the surrounding environment below the vehicle, and some color patterns are not completely removed. This indicates that a small $r$ cannot cover the diversity of artist style characteristics, resulting in incomplete style suppression. Within the dimension range of $\{12,15,18,21\}$, CAPE achieves a better balance between style suppression and content preservation. The brushstroke textures and color patterns annotated in both foreground and background regions are sufficiently suppressed, while structural information is well preserved. When $r$ is further increased, additional textures and artifacts begin to appear in the background and the region below the vehicle, indicating that a larger $r$ introduces excessive noise features and consequently degrades visual quality.

Furthermore, we investigate whether the generalized eigenvalue ranking can effectively select more useful style directions. We divide the direction selection strategies into four groups: Top-k, which selects the first $k$ directions with the largest generalized eigenvalues $[u_1,\ldots,u_k]$; Bottom-k, which selects the last $k$ directions with the smallest generalized eigenvalues $[u_{D-k+1},\ldots,u_D]$; Random-k, which randomly samples $k$ directions from the complete generalized eigenvector set; and Random Subspace, which constructs a random orthogonal subspace with the same dimensionality in the original feature space.

\begin{table}[t]
	\centering
	\caption{Target-style suppression with different direction-selection strategies.}
	\label{tab10}
	\footnotesize
	\setlength{\tabcolsep}{5pt}
	
	\begin{tabular*}{\columnwidth}{@{\extracolsep{\fill}}ccccc}
		\toprule
		
		$k$
		&
		Top-k
		&
		Bottom-k
		&
		Random-k
		&
		Random Subspace
		\\
		
		\midrule
		
		1
		&
		\textbf{-0.001592}
		&
		-0.001611
		&
		-0.001830
		&
		-0.001770
		\\
		
		5
		&
		\textbf{0.008412}
		&
		0.005367
		&
		0.004510
		&
		0.004044
		\\
		
		10
		&
		0.012353
		&
		0.012669
		&
		0.012005
		&
		\textbf{0.012716}
		\\
		
		15
		&
		\textbf{0.021918}
		&
		0.019880
		&
		0.018557
		&
		0.016450
		\\
		
		\bottomrule
		
	\end{tabular*}
\end{table}

We measure the style score difference between the original and edited images, where a larger value indicates stronger target-style suppression. As shown in Table \ref{tab10}, Top-k achieves the strongest style suppression performance at both $k=5$ and $k=15$. Although different strategies produce similar results when $k=10$, the overall results demonstrate that selecting directions corresponding to larger generalized eigenvalues provides more effective and stable target-style suppression.

\subsubsection{Feature Extraction Diffusion Position}

\begin{figure}[h]
	\centering
	\includegraphics[width=1\columnwidth]{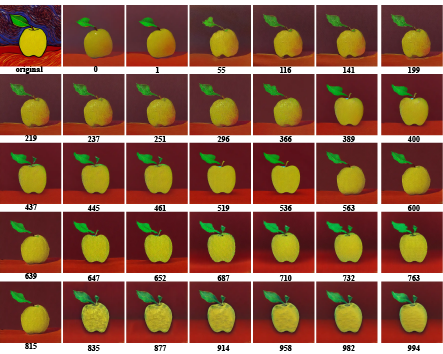} 
	\caption{Qualitative stability of CAPE across feature probe positions. We randomly sample 35 positions from the 1000 candidate scheduler indices and independently estimate the style eigenbasis at each position. CAPE maintains consistent target style suppression over a broad range of probe positions, demonstrating that eigenbasis estimation does not rely on a narrowly selected diffusion position.}
	\label{figure11}
\end{figure}

We further investigate the impact of feature extraction positions on CAPE. We extract intermediate features from different diffusion step intervals while keeping all other parameters unchanged, estimate the corresponding style directions, and perform the same purification process. We randomly sample 35 positions from the 1000 candidate indices defined by the scheduler, as shown in Figure \ref{figure11}. Across a broad diffusion step range (0 ~ 900), CAPE consistently suppresses brushstrokes, textures, and overall color characteristics associated with the target artist style. Meanwhile, it preserves the main object shape, primary contours, and overall composition without showing significant dependence on a specific diffusion position, indicating that the style directions estimated from contrastive triplets remain stable across a wide range of diffusion times.

When the feature extraction position approaches the end of the diffusion process, the target style can still be effectively suppressed. However, some samples begin to exhibit slight local texture anomalies and surface pattern changes. These variations suggest that feature extraction at later stages may cause the estimated style directions to contain more fine-grained appearance information that has already been formed during generation. Overall, CAPE maintains a broad effective range with respect to feature extraction positions and only exhibits slight performance degradation at very late diffusion stages.

\subsubsection{Triplet Reference}

\begin{figure}[h]
	\centering
	\includegraphics[width=1\columnwidth]{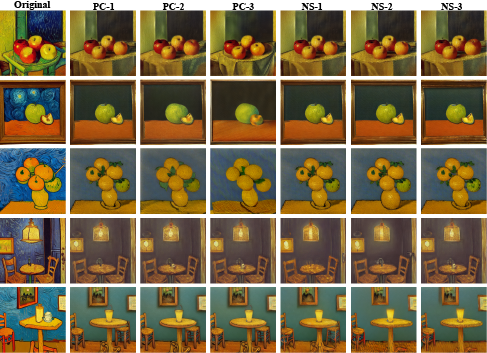} 
	\caption{Triplet reference sensitivity results. CAPE maintains consistent target style suppression and content structure across different positive content and negative style references, showing stable eigenbasis estimation under different reference choices.}
	\label{figure12}
\end{figure}

We further investigate the sensitivity of positive content and negative style selections in triplet construction. We design two groups of prompts. For each anchor, we construct five different positive contents (PC1 to PC5) and five different negative styles (NS1 to NS5). For example, PC1 keeps the Anchor and Negative fixed while replacing the content in Positive. NS1 keeps the Anchor and Positive fixed while replacing the style in Negative. We re-estimate $U_{\mathrm{style}}$ under each setting while keeping other configurations unchanged. The results are shown in Figure \ref{figure12}.

The results show that under different reference choices, CAPE consistently achieves target style suppression and content structure preservation. The generated results do not exhibit visual transfer corresponding to the selected Positive content or Negative style, indicating that eigenbasis estimation does not depend on a specific positive content or negative style reference.

\section{Limitations}

CAPE focuses on artist styles whose visual characteristics are primarily expressed through rendering attributes, including brushstrokes, textures, local color processing, and surface appearance. For styles characterized by object deformation, scene geometry, composition, or distinctive structural layouts, purification may be more challenging when relying only on suppression based on appearance features. Finally, CAPE is an inference-time protection mechanism and does not permanently remove the target concept from model parameters. Therefore, it focuses on controlling style expression during generation rather than preventing the model from learning the concept or achieving irreversible concept deletion.

\section{Conclusion}
\label{sec: Conclusion}
This work revisits the problem of artist style protection against style mimicry in text-to-image diffusion models. Through intervention diagnosis, we find that artist style erasure exhibits broader causal support than ordinary object erasure, while its non-target collateral effects are less restricted to local semantic neighborhoods. These observations indicate that directly treating artist style protection from the perspective of improving erasure strength and editing precision has limitations, motivating us to reformulate artist style protection as style purification. Based on this perspective, we propose the training-free CAPE framework, which identifies cross-content stable target style related directions through contrastive triplets and generalized eigenbasis estimation, and performs adaptive suppression through the Adaptive Suppression Controller on the K/V pathways. Experiments demonstrate that CAPE can effectively suppress target artist brushstrokes, textures, and local rendering characteristics while better preserving major semantic entities and visual structures. The results show that restricting target style expression during generation provides a lightweight and controllable alternative for artist style protection compared with permanent checkpoint erasure.

{
    \small
    \bibliographystyle{ieeenat_fullname}
    \bibliography{main}
}


\clearpage

\setcounter{section}{0}

\renewcommand\thesection{\Alph{section}}
\renewcommand{\theequation}{S\arabic{equation}}
\renewcommand{\thefigure}{S\arabic{figure}}
\renewcommand{\thetable}{S\arabic{table}}

\twocolumn[{
\renewcommand\twocolumn[1][]{#1}
\maketitlesupplementary
\vspace{-0.5cm}
}]

\section{Distributed Causal Support of Erasure}

Algorithms and Models. The experiment uses SPEED \cite{li2026speed}. For each target concept, we first apply SPEED for editing and generate 100 images for CLIP Accuracy evaluation. A target concept recognition accuracy below 3\% is considered successful erasure, and the corresponding edited model $M_{edit}$ is obtained. The complete evaluation set contains 73 artist styles, 100 object concepts, and 100 celebrity identities, as detailed in Table \ref{tab:A1}.

\section{Non-target Collateral Effect}

The non-target collateral experiment contains 50 artist style targets and 50 object targets. For each target, we construct 24 non-target probes, including 8 Near probes, 8 Middle probes, and 8 Far probes. For artist styles, Near probes are selected from the same style family as the target, Middle probes are selected from predefined adjacent style families, and Far probes are selected from non-adjacent style families. For object targets, we construct three levels based on the same principle using a predefined semantic family taxonomy, including same family, adjacent family, and unrelated family. Due to space constraints, in Table \ref{tab:B1}, we have separately presented the non-target concept lists for five artist styles and the ordinary objects.

\section{Cross-architecture Evaluation}
Since FLUX adopts the flow matching training objective and the MMDiT architecture, its generation process is described by a conditional update process along a continuous-time state trajectory, and state updates are completed through a finite number of numerical solution steps, which differs from the discrete denoising update form. Therefore, when migrating CAPE to FLUX, we retain its core idea of estimating target style related directions through conditional contrasts and performing style suppression during inference, while adapting the response representation, subspace estimation method, and intervention location to the architecture.

In FLUX, we adopt the contrastive triplet:
\begin{equation}
	\small
  A=(c_0,s^\star),\quad P=(c_m,s^\star),\quad Z=(c_0,\varnothing)  
\end{equation}
$Z$ denotes the same content prompt after removing the explicit artist style description. Unlike Stable Diffusion, which extracts an intermediate feature representation from a single diffusion state, the generation process of FLUX based on flow matching predicts the generation direction $v_\theta(x_t,t,c)$ through the state $x_t$. Under the same current state $x_t$ and timestep $t$, the image token output responses corresponding to the triplet conditions are computed respectively:
\begin{equation}
	\small
	R_{A,t}=R_A(x_t,t),\qquad R_{P,t}=R_P(x_t,t),\qquad R_{Z,t}=R_Z(x_t,t)  
\end{equation}
On this basis, the style condition difference is defined as:
\begin{equation}
	\small
	D_{s,t}=R_{A,t}-R_{Z,t}
\end{equation}
Since $A$ and $Z$ have the same content $c_0$, but only $A$ contains the target style $s^\star$, $D_{s,t}$ mainly characterizes the response change caused by introducing the target style description under the shared state and fixed content condition. Correspondingly, the content change difference is defined as:
\begin{equation}
	\small
	D_{c,t}=R_{A,t}-R_{P,t}
\end{equation}
Since $A$ and $P$ both contain the same target style $s^\star$ but correspond to different contents $c_0$ and $c_m$, $D_{c,t}$ mainly characterizes the response change caused by content condition changes and serves as content related interference to be suppressed in the generalized eigenvalue problem. Introducing differences in FLUX is used to represent contrastive responses under shared state conditions and to approximately distinguish the target style condition effect from the content change effect. 
Considering that the style features of FLUX are gradually formed along the generation trajectory and that the response distributions at different stages are not completely consistent, we do not use a single global subspace covering all intervention steps, but divide the early inference steps into four non-overlapping temporal windows $W$, , namely $[0,4],\ [5,9],\ [10,14],\ [15,19]$. For each temporal window $W$, we directly use the style condition difference $D_{s,t}$ and content condition difference $D_{c,t}$ within that window to solve the following generalized Rayleigh quotient maximization problem:
\begin{equation}
	\small
	\underset{u\neq 0}{\max}\;\mathcal{J}_s^W(u) = \underset{u\neq 0}{\max}\; \frac{ u^\top \left( \sum\limits_{t\in W} w_tD_{s,t}^{\top}D_{s,t} \right)u }{ u^\top \left[ \sum\limits_{t\in W} w_tD_{c,t}^{\top}D_{c,t} + \varepsilon_W I \right]u }
\end{equation}
Here, $w_t$ denotes the trajectory weight of the window to which timestep $t$ belongs, and $\varepsilon_W$ is the regularization coefficient for ensuring numerical stability. Since $\mathcal{J}_s^W(au)=\mathcal{J}_s^W(u)$ holds for any nonzero constant $a$, the maximization problem can therefore be solved under the following normalization constraint:
\begin{equation}
	\footnotesize
	u^\top
	\left[
	\sum\limits_{t\in W}
	w_tD_{c,t}^{\top}D_{c,t}
	+
	\varepsilon_W I
	\right]u
	=1
\end{equation}
We use the Lagrange multiplier method to solve this constrained optimization problem and construct the corresponding Lagrangian function:
\begin{equation}
	\footnotesize
	\begin{aligned}
		\mathcal{L}_W(u,\rho)
		={}&
		u^\top
		\left(
		\sum\limits_{t\in W}
		w_tD_{s,t}^{\top}D_{s,t}
		\right)u
		\\
		&-
		\rho
		\left(
		u^\top
		\left[
		\sum\limits_{t\in W}
		w_tD_{c,t}^{\top}D_{c,t}
		+
		\varepsilon_W I
		\right]u
		-1
		\right)
	\end{aligned}
\end{equation}
Setting the gradient of $\mathcal{L}_W(u,\rho)$ with respect to $u$ to zero gives the first-order optimality condition:
\begin{equation}
	\footnotesize
	\begin{aligned}
		\nabla_u\mathcal{L}_W
		={}&
		2\left(
		\sum\limits_{t\in W}
		w_tD_{s,t}^{\top}D_{s,t}
		\right)u
		\\
		&-
		2\rho\left[
		\sum\limits_{t\in W}
		w_tD_{c,t}^{\top}D_{c,t}
		+
		\varepsilon_W I
		\right]u
		=0
	\end{aligned}
\end{equation}
This further yields the following generalized eigenvalue problem:
\begin{equation}
	\small
	\left( \sum\limits_{t\in W}w_tD_{s,t}^{\top}D_{s,t} \right)u = \rho \left[ \sum\limits_{t\in W}w_tD_{c,t}^{\top}D_{c,t} + \varepsilon_W I \right]u
\end{equation}
By solving the above generalized eigenvalue problem, the first $r$ directions are selected according to the generalized eigenvalue $\rho$ in descending order and orthogonalized to obtain the style related subspace corresponding to window $W$:
\begin{equation}
	\small
	U_{\mathrm{style}}^{W}=\left[u_1^W,u_2^W,\ldots,u_r^W\right]  
\end{equation}
During inference, for timestep $t$ belonging to temporal window $W$, the style condition difference is first recomputed at the current generation state $x_t$:
\begin{equation}
	\small
	D_{s,t}=R_{A,t}-R_{Z,t}  
\end{equation}
The style condition difference is then projected token by token onto the style related subspace corresponding to window $W$:
\begin{equation}
	\small
	D_{s,t}^{\mathrm{proj},W} = D_{s,t} U_{\mathrm{style}}^{W} \left(U_{\mathrm{style}}^{W}\right)^\top  
\end{equation}
To avoid losing effective suppression components in the original style condition difference by using only the low-rank projection, we adopt the residual mixing form:
\begin{equation}
	\small
	D_t = \eta D_{s,t}^{\mathrm{proj},W} + (1-\eta)D_{s,t}  
\end{equation}
Here, $\eta$ controls the contribution of the style subspace. The image branch response is then adjusted as:
\begin{equation}
	\small
	R_{A,t}^{\prime} = R_{A,t} - \beta_tD_t  
\end{equation}
We use 50 inference steps to generate images and perform response intervention only at steps 0–19. The window weights are $w=0.5,1.0,1.5,1.2$, the rank is $r=32$, and the projection mixing coefficient is $\eta=0.35$. All results are generated using the same prompts and random seeds as the original FLUX outputs.

Although FLUX differs significantly from Stable Diffusion in both its training objective and network architecture, the adapted CAPE mechanism can still suppress style related patterns such as swirling starry sky brushstrokes, dense painterly textures, and highly saturated color transitions, while preserving the main scene layout and object structure.

\begin{table*}[t]
	\centering
	\caption{Target concepts used in the distributed causal support analysis.}
	\label{tab:A1}
	
	\small
	\renewcommand{\arraystretch}{1.2}
	
	\begin{tabularx}{\textwidth}{>{\bfseries}p{2.2cm}X}
		
		\toprule
		
		Category & Concept \\
		
		\midrule

		Artist style &
		Agnes Martin, Albert Bierstadt, Alphonse Mucha, Amedeo Modigliani, Andre Derain, Aubrey Beardsley, Caspar David Friedrich, Chaim Soutine, Clyfford Still, Dali, David Hockney, Dorothea Tanning, Edgar Degas, Edvard Munch, Egon Schiele, El Greco, El Lissitzky, Ernst Ludwig Kirchner, Fernand Leger, Fra Angelico, Francis Bacon, Francis Picabia, Francisco Goya, Franz Kline, Georges Braque, Gerhard Richter, Giorgio de Chirico, Gustave Courbet, Henri Matisse, Henri Rousseau, Henri de Toulouse-Lautrec, Hieronymus Bosch, Hilma af Klint, Jean Arp, Jean-Francois Millet, Jean-Honore Fragonard, Joan Miro, Johannes Vermeer, John Constable, John Everett Millais, Juan Gris, Kandinsky, Kazimir Malevich, Keith Haring, Leonardo da Vinci, Leonora Carrington, Man Ray, Marcel Duchamp, Mary Cassatt, Max Ernst, Monet, Odilon Redon, Paul Cezanne, Paul Gauguin, Paul Klee, Picasso, Piet Mondrian, Pieter Bruegel the Elder, Rembrandt, Remedios Varo, Rene Magritte, Robert Delaunay, Roy Lichtenstein, Sonia Delaunay, Theo van Doesburg, Thomas Gainsborough, Van Gogh, Warhol, William Blake, William-Adolphe Bouguereau, Winslow Homer, Yayoi Kusama, Yves Tanguy
		\\[3pt]
		
		\midrule
		
		Object &
		Mickey Mouse, Naruto, Spider-Man, airplane, ambulance, apple, backpack, banana, bed, bicycle, bird, boat, book, bottle, bowl, bridge, broccoli, bus, butterfly, cake, camera, car, carrot, castle, cat, chair, clock, clothing, comb, corgi, cup, dog, doughnut, drum, eiffel tower, elephant, fire truck, fish, flower, fork, frog, fruit, garbage truck, giraffe, guitar, hamburger, hammer, hat, helmet, horse, ice cream, knife, lamp, laptop, lemon, lesser panda, machine, mango, motorcycle, orange, pear, phone, piano, pineapple, pizza, plate, pyramid, rabbit, rose, sandwich, saxophone, scissors, screwdriver, shark, shoe, snake, sneakers, sofa, spoon, squirrel, statue, strawberry, table, taxi, television, tomato, toothbrush, towel, traffic light, train, tree, truck, trumpet, turtle, umbrella, vegetable, violin, watermelon, wrench, zebra
		\\[3pt]
		
		\midrule
		
		Celebrity &
		50 Cent, Aaron Paul, Adam Driver, Adam Sandler, Adele, Al Gore, Al Pacino, Alan Rickman, Alicia Keys, Amy Adams, Angelina Jolie, Anne Hathaway, Ariana Grande, Barack Obama, Ben Affleck, Benedict Cumberbatch, Bernie Sanders, Beyoncé, Bill Clinton, Brad Pitt, Bradley Cooper, Britney Spears, Bruce Willis, Bruno Mars, Bryan Cranston, Cameron Diaz, Celine Dion, Charlize Theron, Chris Evans, Chris Pratt, Christian Bale, Cristiano Ronaldo, David Beckham, David Bowie, Denzel Washington, Donald Trump, Drake, Drew Barrymore, Dua Lipa, Dwayne Johnson, Ed Sheeran, Elon Musk, Elton John, Emma Stone, Emma Watson, George Clooney, Halle Berry, Harrison Ford, Hillary Clinton, Hugh Jackman, Idris Elba, Jack Black, Jack Nicholson, Jackie Chan, Jake Gyllenhaal, Jamie Foxx, Jay Z, Jennifer Aniston, Jennifer Lawrence, Jennifer Lopez, Jim Carrey, Joaquin Phoenix, Joe Biden, Johnny Depp, Julia Roberts, Justin Bieber, Katy Perry, Keanu Reeves, Kevin Hart, Kim Kardashian, Kobe Bryant, Lady Gaga, Leonardo DiCaprio, Lionel Messi, Margot Robbie, Matt Damon, Matthew McConaughey, Miley Cyrus, Morgan Freeman, Natalie Portman, Nicole Kidman, Oprah Winfrey, Reese Witherspoon, Rihanna, Robert Downey Jr, Roger Federer, Sandra Bullock, Scarlett Johansson, Selena Gomez, Serena Williams, Shakira, Sofia Vergara, Stephen Curry, Steve Jobs, Taylor Swift, Vladimir Putin, Whoopi Goldberg, Will Smith, Zendaya, Zoe Saldana
		\\
		
		\bottomrule
		
	\end{tabularx}
	
\end{table*}

\begin{table*}[t]
	\centering
	\caption{Target concepts and corresponding Near, Middle, and Far non target probes.}
	\label{tab:B1}
	\footnotesize
	\setlength{\tabcolsep}{3pt}
	\renewcommand{\arraystretch}{1.15}
	\begin{tabularx}{\textwidth}{@{}c p{2.3cm} >{\raggedright\arraybackslash}X >{\raggedright\arraybackslash}X >{\raggedright\arraybackslash}X@{}}
		\toprule
		\textbf{Category} & \textbf{target concept} & \textbf{Near 8} & \textbf{Middle 8} & \textbf{Far 8} \\
		\midrule
		
		\multirow{5}{*}{\textbf{artist}} 
		& Alphonse Mucha & organic arabesque composition; vintage theatrical poster; symbolist decorative illustration; ornate bookplate design; ornamental art nouveau illustration; flowing botanical linework; decorative belle epoque poster; whiplash curve design & dreamlike religious illustration; spiritual allegorical image; hand-colored visionary print; cosmic poetic scene; symbolic gothic watercolor; visionary romantic illustration; mystical literary watercolor; symbolic biblical drawing & mass-media appropriation art; vivid product illustration; repeated icon composition; retro pop poster; hard-edged popular imagery; pop art screenprint; bold commercial graphic painting; comic-inspired modern artwork \\
		\cmidrule(l){2-5}
		& Amedeo Modigliani & linear figurative painting; restrained expressionist portrait; simplified seated figure; earth-toned studio portrait; modernist nude study; elongated modernist portrait; simplified sculptural face; muted bohemian portrait painting & organic arabesque composition; vintage theatrical poster; symbolist decorative illustration; ornate bookplate design; ornamental art nouveau illustration; flowing botanical linework; decorative belle epoque poster; whiplash curve design & dynamic color-wheel composition; musical cubist color field; radiant modernist geometry; interlocking spectral disks; abstract urban rhythm; vibrating concentric forms; color-based cubist painting; orphic cubist abstraction \\
		\cmidrule(l){2-5}
		& Andre Derain & loose colorful figure painting; radiant decorative landscape; fauvist color painting; vivid non-naturalistic landscape; bold chromatic portrait; high-saturation brushwork; decorative fauvist interior; expressive color-block painting & distorted emotional portrait; agitated brushwork composition; intense psychological landscape; raw chromatic painting; angular urban expressionism; anxious modernist scene; rough impasto portrait; emotionally charged color field & dreamlike religious illustration; spiritual allegorical image; hand-colored visionary print; cosmic poetic scene; symbolic gothic watercolor; visionary romantic illustration; mystical literary watercolor; symbolic biblical drawing \\
		\cmidrule(l){2-5}
		& Caspar David Friedrich & misty mountain painting; luminous pastoral landscape; stormy romantic scene; moonlit wilderness painting; heroic natural panorama; poetic rural landscape; cloud-filled coastal painting; picturesque countryside scene & ornate narrative oil painting; poetic Shakespearean scene; crisp decorative realism; pre-raphaelite painting; jewel-toned medieval portrait; highly detailed romantic figure; botanical literary illustration; luminous Victorian narrative scene & orphic cubist abstraction; simultaneous color circles; rhythmic prismatic painting; luminous geometric color; circular chromatic abstraction; dynamic color-wheel composition; musical cubist color field; radiant modernist geometry \\
		\cmidrule(l){2-5}
		& Edvard Munch & visceral figurative painting; expressionist figure painting; distorted emotional portrait; agitated brushwork composition; intense psychological landscape; raw chromatic painting; angular urban expressionism; anxious modernist scene & vivid non-naturalistic landscape; bold chromatic portrait; high-saturation brushwork; decorative fauvist interior; expressive color-block painting; bright Mediterranean scene; flat intense color composition; wild-beast modern painting & spiritual abstract painting; mystical geometric symbolism; esoteric color diagram; theosophical abstraction; cosmic circular composition; visionary nonobjective painting; symbolic spectral geometry; meditative abstract cosmology \\
		
		\midrule
		
		\multirow{6}{*}{\textbf{object}} 
		& airplane & hot air balloon; drone; seaplane; biplane; airship; fighter jet; cargo plane; paraglider & construction site; airport terminal; harbor; subway station; warehouse; service road; public square; city street & laundry basket; mirror; rug; soap dispenser; vacuum cleaner; iron; doormat; pillow \\
		\cmidrule(l){2-5}
		& apple & papaya; fig; apricot; lime; coconut; dragon fruit; guava; orange & vase; kettle; plate; mug; teapot; glass; tray; pitcher & grasshopper; ladybug; ant; praying mantis; cicada; wasp; firefly; caterpillar \\
		\cmidrule(l){2-5}
		& castle & museum building; lighthouse; stadium; monastery; palace; cathedral; tower; fortress & bedroom; bathroom; dining room; office desk; laundry room; garden; workshop; classroom & firefly; caterpillar; bee; dragonfly; beetle; moth; grasshopper; ladybug \\
		\cmidrule(l){2-5}
		& corgi & beagle; poodle; husky; bulldog; golden retriever; border collie; terrier; shiba inu & camel; rhinoceros; hippopotamus; donkey; moose; fox; otter; squirrel & hammer; screwdriver; pliers; drill; saw; chisel; tape measure; socket wrench \\
		\cmidrule(l){2-5}
		& guitar & harp; accordion; tambourine; xylophone; oboe; banjo; harmonica; double bass & library; studio; kitchen; bedroom; bathroom; office desk; garden; workshop & speedboat; gondola; catamaran; ship; canoe; kayak; sailboat; ferry \\
		\cmidrule(l){2-5}
		& laptop & game console; headphones; printer; smartwatch; smartphone; tablet computer; desktop monitor; keyboard & magazine; notebook; map; poster; letter; calendar; comic book; dictionary & pitcher; saucer; jar; food container; serving dish; vase; kettle; plate \\
		\bottomrule
	\end{tabularx}
\end{table*}

\section{Sensitivity Analysis}

\subsection{Feature Extraction Diffusion Position}

\begin{figure*}[h] 
	\centering 
	\includegraphics[width=1\textwidth]{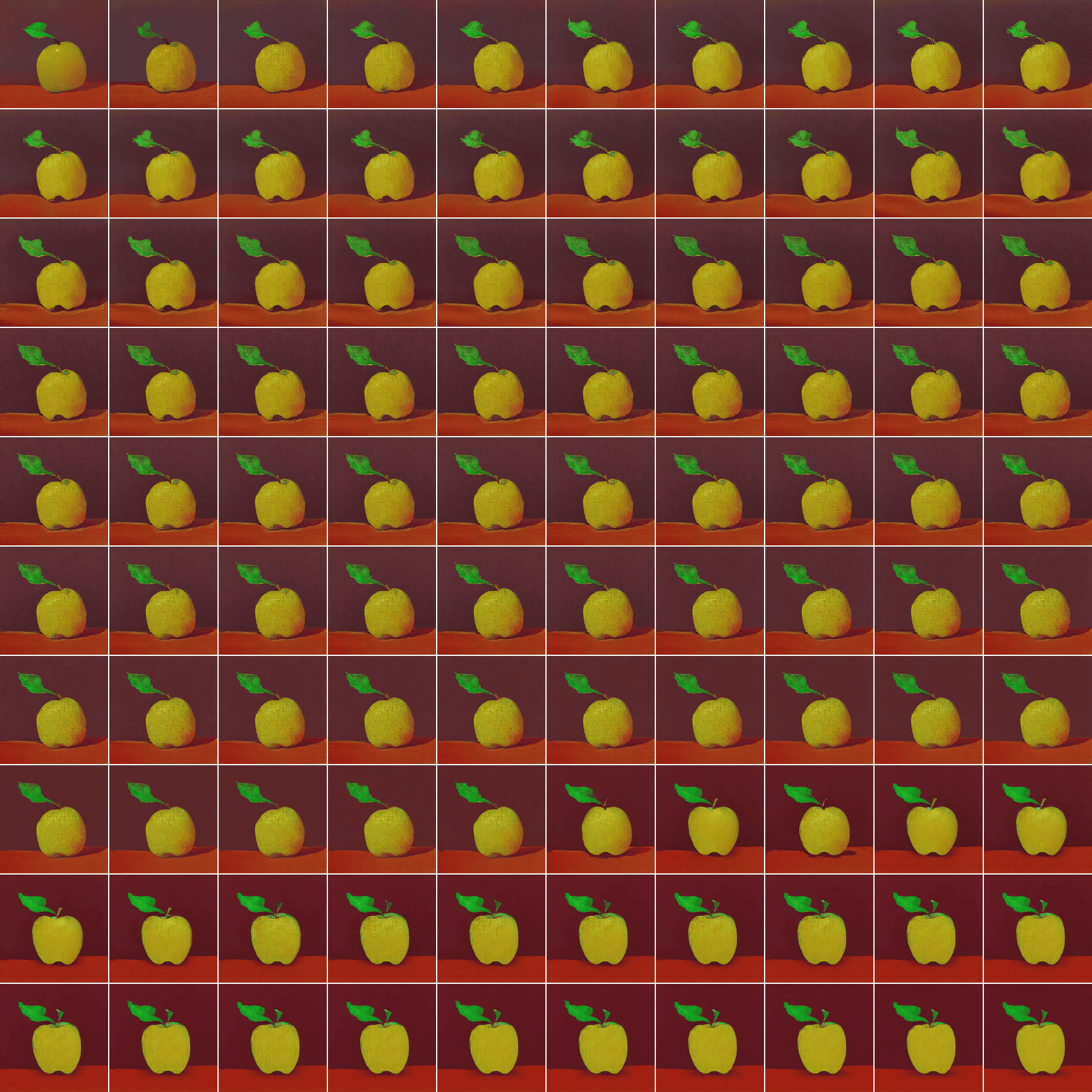} 
	\caption{Results over 1000 diffusion steps with an interval of 5 (steps=5,15,20,...,495).} 
	\label{figureA1} 
\end{figure*}

\begin{figure*}[h] 
	\centering 
	\includegraphics[width=1\textwidth]{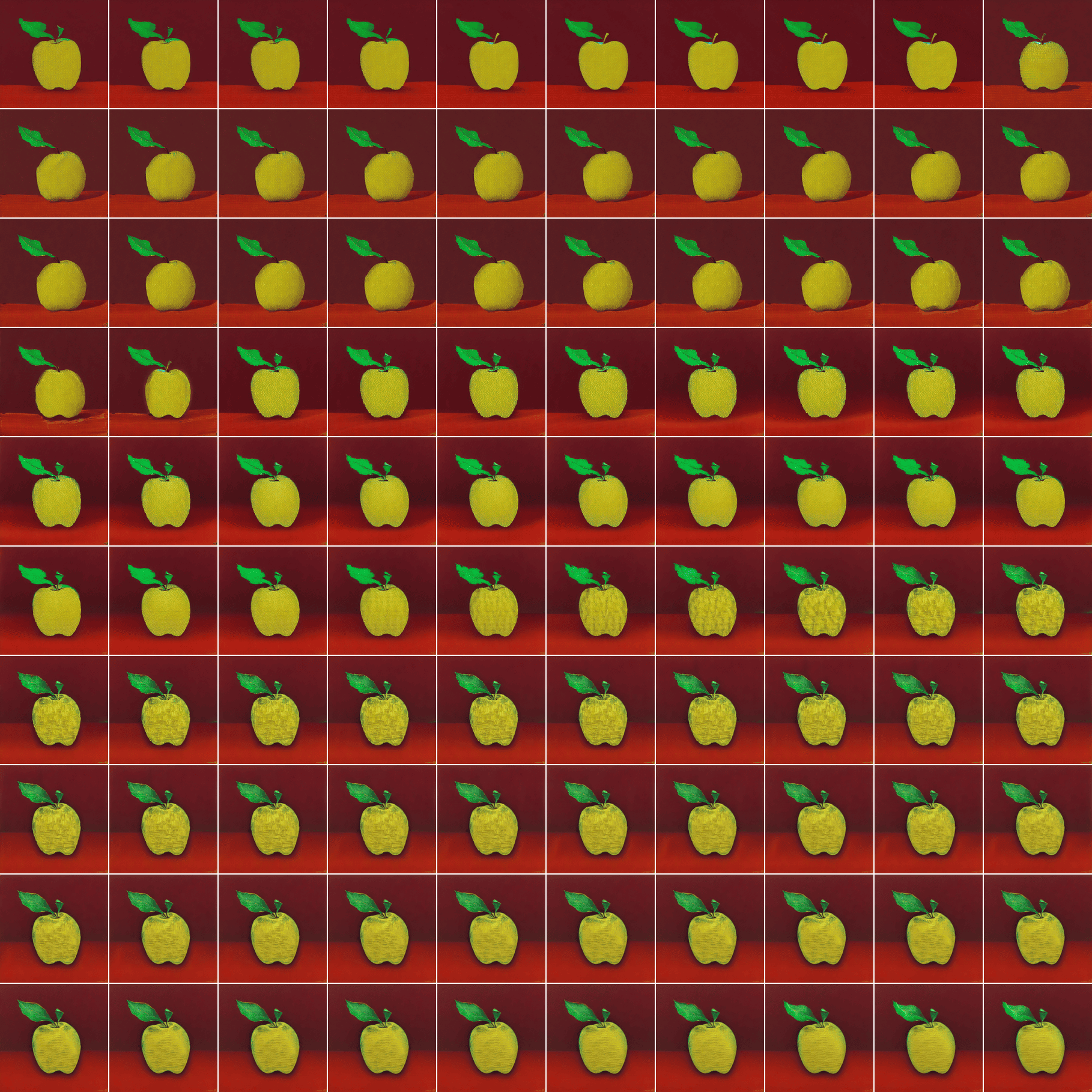} 
	\caption{Results over 1000 diffusion steps with an interval of 5 (steps=500,505,...,995).} 
	\label{figureA2} 
\end{figure*}

We present the complete results of the 1000 diffusion steps with an interval of 5 steps (Figure \ref{figureA1},\ref{figureA2}), as shown in Figures \ref{figureA1},\ref{figureA2} and Table \ref{tab:C1}.

\subsection{Triplet Reference}
\begin{figure*}[h] 
	\centering 
	\includegraphics[width=1\textwidth]{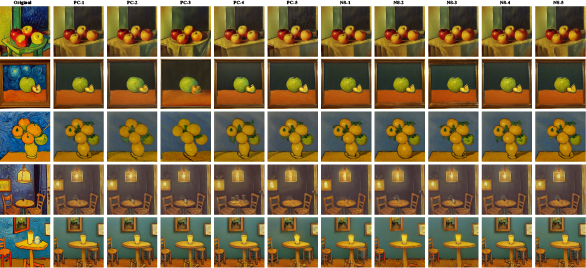} 
	\caption{Results under 10 different reference settings.} 
	\label{figureA3} 
\end{figure*}

In Figure \ref{figureA3}, we present the visual results of all 10 groups and list the corresponding triplet settings.

\begin{table*}[t]
	\centering
	\caption{Triplet reference prompts used in the sensitivity analysis.}
	\label{tab:C1}
	\small
	\setlength{\tabcolsep}{4pt}
	\renewcommand{\arraystretch}{1.05}
	
	\begin{tabularx}{\textwidth}{@{}c c c X@{}}
		\toprule
		\textbf{Prompt} & \textbf{Type} & \textbf{ID} & \textbf{Prompt Description} \\
		\midrule

		
		\multirow{11}{*}{P1}
		& Anchor & A &
		A still life of several apples beside a ceramic jug on a wooden table, with a folded cloth and a simple wall behind the objects in the style of Vincent van Gogh.
		\\
		\cmidrule(l){2-4}
		
		& \multirow{5}{*}{Positive}
		& P1 &
		A bright coastal harbor with small sailboats and a stone quay in the style of Vincent van Gogh.
		\\
		& & P2 &
		A cypress-lined country road leading toward distant hills in the style of Vincent van Gogh.
		\\
		& & P3 &
		A wheat field beside a small farmhouse under an open sky in the style of Vincent van Gogh.
		\\
		& & P4 &
		A flowered garden terrace with a table and wicker chairs in the style of Vincent van Gogh.
		\\
		& & P5 &
		A quiet riverside village with a stone bridge and leafy trees in the style of Vincent van Gogh.
		\\
		
		\cmidrule(l){2-4}
		
		& \multirow{5}{*}{Negative}
		& N1 &
		A still life of several apples beside a ceramic jug on a wooden table, with a folded cloth and a simple wall behind the objects in the style of Maurice de Vlaminck.
		\\
		& & N2 &
		A still life of several apples beside a ceramic jug on a wooden table, with a folded cloth and a simple wall behind the objects in the style of Chaïm Soutine.
		\\
		& & N3 &
		A still life of several apples beside a ceramic jug on a wooden table, with a folded cloth and a simple wall behind the objects in the style of Abram Efimovich Arkhipov.
		\\
		& & N4 &
		A still life of several apples beside a ceramic jug on a wooden table, with a folded cloth and a simple wall behind the objects in the style of Childe Hassam.
		\\
		& & N5 &
		A still life of several apples beside a ceramic jug on a wooden table, with a folded cloth and a simple wall behind the objects in the style of Charles Camoin.
		\\

		\midrule

		
		\multirow{11}{*}{P2}
		& Anchor & A &
		A modest cafe interior with wooden chairs, a few small tables, a hanging lamp, and a doorway opening toward a quiet street in the style of Vincent van Gogh.
		\\
		\cmidrule(l){2-4}
		
		& \multirow{5}{*}{Positive}
		& P1 &
		A still life of apples and a ceramic jug on a plain wooden table in the style of Vincent van Gogh.
		\\
		& & P2 &
		A cypress-lined country road leading toward distant hills in the style of Vincent van Gogh.
		\\
		& & P3 &
		A wheat field beside a small farmhouse under an open sky in the style of Vincent van Gogh.
		\\
		& & P4 &
		A bright coastal harbor with small sailboats and a stone quay in the style of Vincent van Gogh.
		\\
		& & P5 &
		A flowered garden terrace with a table and wicker chairs in the style of Vincent van Gogh.
		\\
		
		\cmidrule(l){2-4}
		
		& \multirow{5}{*}{Negative}
		& N1 &
		A modest cafe interior with wooden chairs, a few small tables, a hanging lamp, and a doorway opening toward a quiet street in the style of Maurice de Vlaminck.
		\\
		& & N2 &
		A modest cafe interior with wooden chairs, a few small tables, a hanging lamp, and a doorway opening toward a quiet street in the style of Chaïm Soutine.
		\\
		& & N3 &
		A modest cafe interior with wooden chairs, a few small tables, a hanging lamp, and a doorway opening toward a quiet street in the style of Abram Efimovich Arkhipov.
		\\
		& & N4 &
		A modest cafe interior with wooden chairs, a few small tables, a hanging lamp, and a doorway opening toward a quiet street in the style of Childe Hassam.
		\\
		& & N5 &
		A modest cafe interior with wooden chairs, a few small tables, a hanging lamp, and a doorway opening toward a quiet street in the style of Charles Camoin.
		\\
		
		\bottomrule
	\end{tabularx}
\end{table*}

\end{document}